\documentclass[10pt,twocolumn,letterpaper]{article}

\usepackage[pagenumbers]{cvpr} 

\usepackage{graphicx}
\usepackage{amsmath}
\usepackage{amssymb}
\usepackage{booktabs}
\usepackage{adjustbox}
\usepackage{pifont}
\usepackage{lipsum}
\usepackage{subcaption}
\usepackage{listings}
\usepackage{xcolor}
\usepackage{algorithm}
\usepackage{algpseudocode}
\usepackage{pbox}
\usepackage{xcolor, colortbl}
\usepackage{multirow}
\definecolor{lightgreen}{cmyk}{0.20,0.02,0.22,0.0}

\usepackage[pagebackref,breaklinks,colorlinks]{hyperref}

\usepackage[capitalize]{cleveref}
\crefname{section}{Sec.}{Secs.}
\Crefname{section}{Section}{Sections}
\Crefname{table}{Table}{Tables}
\crefname{table}{Tab.}{Tabs.}

\def\cvprPaperID{3964} 
\def\confName{CVPR}
\def\confYear{2022}

\begin{document}

\title{Learned Queries for Efficient Local Attention}

\author{Moab Arar\\
Tel-Aviv University\\
\and
Ariel Shamir\\
Reichman University\\
\and
Amit H. Bermano\\
Tel-Aviv University
}


\maketitle
\begin{abstract}
Vision Transformers (ViT) serve as powerful vision models. Unlike convolutional neural networks, which dominated vision research in previous years, vision transformers enjoy the ability to capture long-range dependencies in the data.
Nonetheless, an integral part of any transformer architecture, the self-attention mechanism, suffers from high latency and inefficient memory utilization, making it less suitable for high-resolution input images. To alleviate these shortcomings, hierarchical vision models locally employ self-attention on non-interleaving windows. This relaxation reduces the complexity to be linear in the input size; however, it limits the cross-window interaction, hurting the model performance. In this paper, we propose a new shift-invariant local attention layer, called query and attend (QnA), that aggregates the input locally in an overlapping manner, much like convolutions. The key idea behind QnA is to introduce learned queries, which allow fast and efficient implementation. We verify the effectiveness of our layer by incorporating it into a hierarchical vision transformer model. We show improvements in speed and memory complexity while achieving comparable accuracy with state-of-the-art models. Finally, our layer scales especially well with window size, requiring up-to x10 less memory while being up-to x5 faster than existing methods. The code is publicly available at~\url{https://github.com/moabarar/qna}.
\end{abstract}
\section{Introduction}
\label{sec:intro}
\begin{figure}
    \centering
    \includegraphics[width=\linewidth]{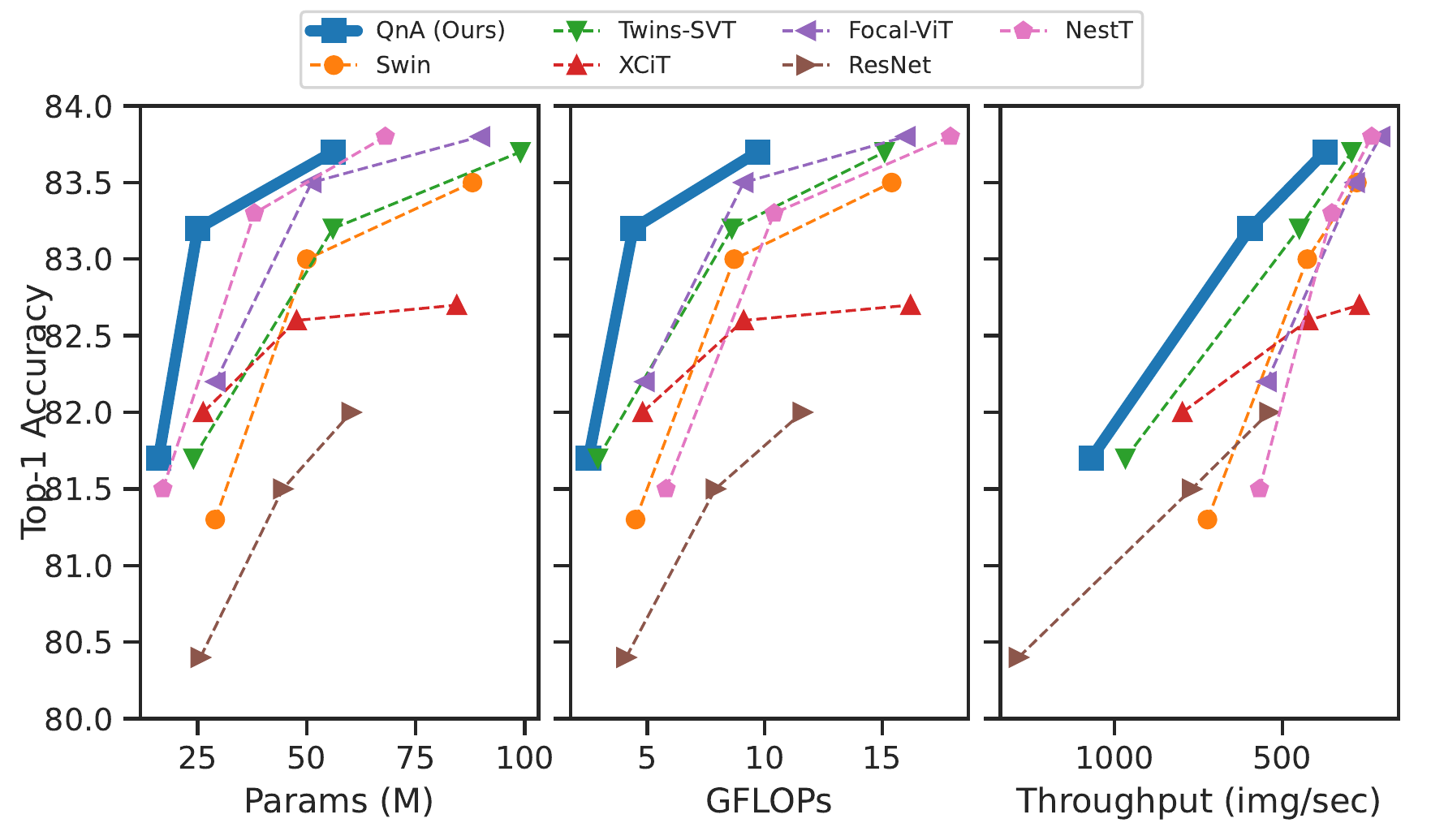}
    \caption{\textbf{Performance-Efficiency Comparisons On $224^2$ Input Size.} QnA-ViT (our method) demonstrates better accuracy-efficiency trade-off compared to state-of-the-art baselines. As suggested by Dehghani et al.~\cite{EfficiencyCorrectReport}, we report the ImageNet-1k~\cite{ImageNet} Top-1 accuracy (y-axis) trade-off with respect to parameter count (left), floating point operations (middle) and inference throughput (right). The throughput is measured using the timm~\cite{timm} library, as tested on NVIDIA V100 with 16GB memory. Other metrics, are from the original publications~\cite{Swin, Twins, FocalViT, NestT, ResNetStrikeBack, XCiT}}
    \label{fig:teaser}
\end{figure}
Two key players take the stage when considering data aggregation mechanisms for image processing. Convolutions were the immediate option of choice. They provide \textit{locality}, which is an established prior for image processing, and \textit{efficiency} while doing so. Nevertheless, convolutions capture local patterns, and extending them to global context is difficult if not impractical. Attention-based models~\cite{AttentionIsAlYouNeed}, on the other hand, offer an adaptive aggregation mechanism, where the aggregation scheme itself is input-dependent, or \textit{spatially dynamic}. These models~\cite{BERT, GPT3} are the \textit{de-facto} choice in the natural-language processing field and have recently blossomed for vision tasks as well.

Earlier variants of the Vision Transformers (ViT)~\cite{ViT} provide global context by processing non-interleaving image patches as word tokens. For these models to be effective, they usually require a vast amount of data~\cite{ViT, JFT300M}, heavy regularization~\cite{DEiT, HowToTrainVit} or modified optimization objectives~\cite{ViTSAM, SAM}. Even more so, it was observed that large scale-training drives the models to attend locally~\cite{CNNvsVIT}, especially for early layers, encouraging the notion that locality is a strong prior.

Local attention mechanisms are the current method of choice for better vision backbones. These backbones follow a pyramid structure similar to convolutional neural networks (CNNs)~\cite{PVT, Twins, MViT, NestT}, and process high-resolution inputs by restricting the self-attention to smaller windows, preferably with some overlap~\cite{HaloNet} or other forms of inter-communication~\cite{Swin, FocalViT, Twins}. The latter approaches naturally induce locality while benefiting from spatially dynamic aggregation. On the other hand, these architectures come at the cost of computational overhead and, more importantly, are not shift-equivariant.

In this paper, we revisit the design of local attention and introduce a new aggregation layer called \textit{Query and Attend} (QnA).  The key idea is to leverage the locality and shift-invariance of convolutions and the expressive power of attention mechanisms. 

In local self-attention, attention scores are computed between all elements comprising the window. This is a costly operation of quadratic complexity in the window size. We propose using learned queries to compute the aggregation weights, allowing linear memory complexity, regardless of the chosen window size. Our layer is also flexible, showing that it can serve as an effective up- or down-sampling operation. We Further observe that combining different queries allows capturing richer feature subspaces with minimal computational overhead. We conclude that QnA layers interleaved with vanilla transformer blocks form a family of hierarchical ViTs that achieve comparable or better accuracy compared to SOTA models while benefiting from up-to x2 higher throughput and fewer parameters and floating-point operations (see Figure~\ref{fig:teaser}).

Through rigorous experiments, we demonstrate that our novel aggregation layer holds the following benefits:
\begin{itemize}
    \item QnA imposes locality, granting efficiency without compromising accuracy.
    \item QnA can serve as a general-purpose layer. For example, strided QnA allows effective down-sampling, and multiple-queries can be used for effective up-sampling, demonstrating improvements over alternative baselines.
    \item QnA naturally incorporates locality into existing transformer-based frameworks. For example, we demonstrate how replacing self-attention layers with QnA ones
    in an attention-based object-detection framework~\cite{DETR} is beneficial for precision, and in particular for small-scale objects. 
\end{itemize}
\section{Related Work}
\label{sec:related}

\paragraph{Convolutional Networks:} CNN-based networks have dominated the computer vision world. For several years now, the computer vision community is making substantial improvements by designing powerful architectures~\cite{VGG, InceptionNet, ResNet, DenseNet, SqueezeAndExcite, ResNext, PyramidNet, EfficientNet, RegNetY}.  A particularly related CNN-based work is RedNet~\cite{Involution}, which introduces an involution operation. This operation extracts convolution kernels for every pixel through linear projection, enabling adaptive convolution operations. Despite its adaptive property, RedNet uses linear projections that lack the expressiveness of the self-attention mechanism.

\paragraph{Vision-Transformers:} The adaptation of self-attention showed promising results in various vision tasks including image recognition~\cite{SASA, AttentionAugmentedConv, ExploringSelfAttention}, image generation~\cite{SAGAN, ImageTransformer}, object-detection~\cite{DualAttention, EmpericalStudy} and semantic-segmentation~\cite{AxialSASA, DualAttention, CCNet}. These models, however, did not place pure self-attention as a dominant tool for vision models. In contrast, vision transformers~\cite{ViT, DEiT}, brought upon a conceptual shift. Initially designed for image classification, these models use global self-attention on \textit{tokenized} image-patches, where each token attends all others.  T2T-ViT~\cite{T2T} further improves the tokenization process via light-weight self-attention at early layers. Similarly, carefully designing a Conv-based STEM-block~\cite{EarlyConv} improves convergence rate and accuracy. CrossViT~\cite{CrossVit} propose processing at both a coarse- and fine-grained patch levels. TNT-ViT~\cite{TNT} on the other hand, splits coarse-patches into locally attending parts. This information is then fused into global attention between patches. ConViT~\cite{ConViT} improves performance by carefully initializing the self-attention block to encourage locality. LeViT~\cite{LeViT} offers an efficient vision transformer through careful design, that combines convolutions and extreme down-sampling. Common to all these models is that, due to memory considerations, expressive feature maps are extracted on very low resolutions, which is not favorable in down-stream tasks such as object-detection.

\paragraph{Local Self-Attention: } Dense prediction tasks involve processing high-resolution images. Global attention is not tractable in this setting, due to quadratic memory and computational requirements. Instead, pyramid architectures employing local attention are used~\cite{Twins, Swin, NestT, HaloNet, CVT, FocalViT}. Typically for such approaches, self-attention is performed within each window, with down-sampling usually applied for global context. Liu et al.~\cite{Swin} propose shifted windows, showing that communication between windows is preferable to independent ones~\cite{PVT}. Halo-Net~\cite{HaloNet} expands the neighborhood of each window to increase context and inter-window communication. Chu et al. ~\cite{Twins} use two-stage self-attention. In the first stage, local attention is employed, while in the second stage a global-self attention is applied on sub-sampled windows. These models however, are not shift-invariant, which is a property we maintain. Closest to our work, is the stand-alone self-attention layer (SASA)~\cite{SASA}. As detailed in later sections, this layer imposes restrictive memory overhead, and is significantly slower, with similar accuracy compared to ours (see~\cref{fig:complexity_comparison}).

\paragraph{Learned Queries:} The concept of learned queries has been explored in the literature in other settings~\cite{MAP, Perceiver, PerceiverIO, goyal2022coordination}. In Set Transformers~\cite{MAP}, learned queries are used to project the input dimension to a smaller output dimension, either for computation consideration or decoding the output prediction. Similarly, the Perceiver networks family~\cite{Perceiver, PerceiverIO} use small latent arrays to encode information from the input array. Goyal et al.~\cite{goyal2022coordination} propose a modification for transformer architectures where learned queries (shared workspace) serve as communication-channel between tokens, avoiding quadratic, pair-wise communication. Unlike QnA, the methods above use cross-attention on the whole input sequence. In QnA, the learned queries are shared across overlapping windows. The information is aggregated locally, leveraging the powerful locality priors that have been so well established through the vast usage of convolutions.
\begin{figure*}
    \centering
    \begin{subfigure}{0.48\textwidth}
        \includegraphics[height=2.1cm]{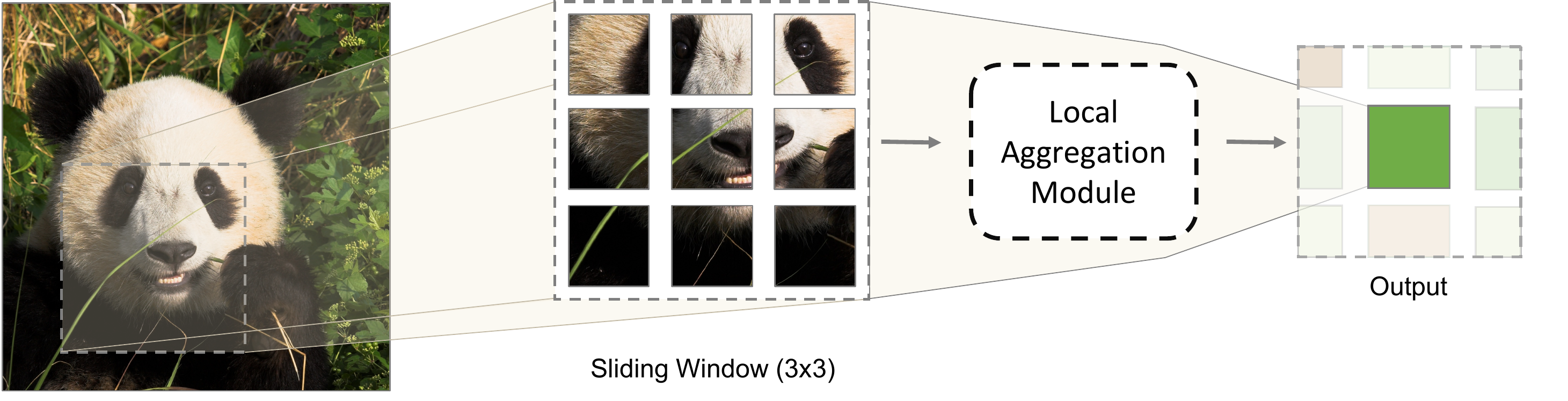}
    \end{subfigure}
     \begin{subfigure}{0.15\textwidth}
        \includegraphics[height=3.75cm]{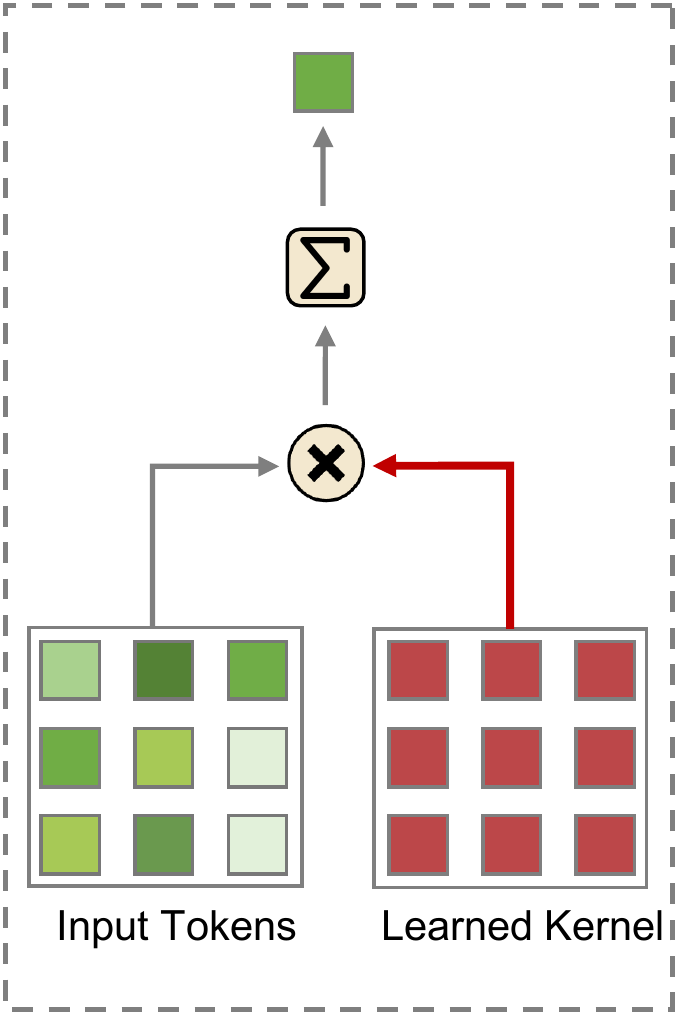}
        \subcaption{Convolution}
        \label{subfig:overview_conv}
    \end{subfigure}
     \begin{subfigure}{0.15\textwidth}
        \includegraphics[height=3.75cm]{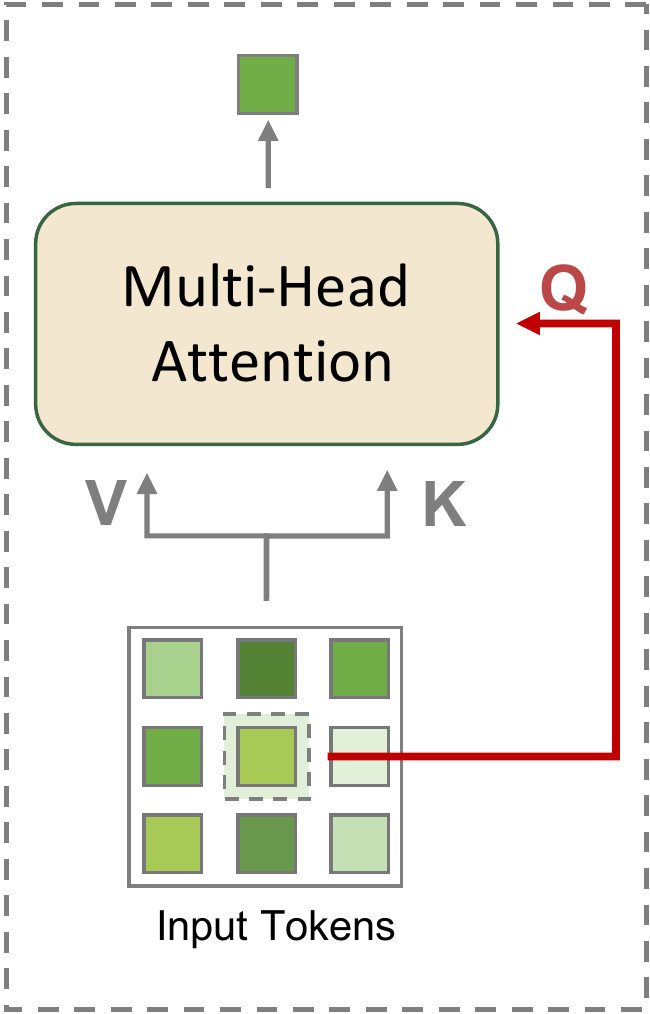}
        \subcaption{SASA~\cite{SASA}}
        \label{subfig:overview_sasa}
    \end{subfigure}
     \begin{subfigure}{0.17\textwidth}
        \includegraphics[height=3.75cm]{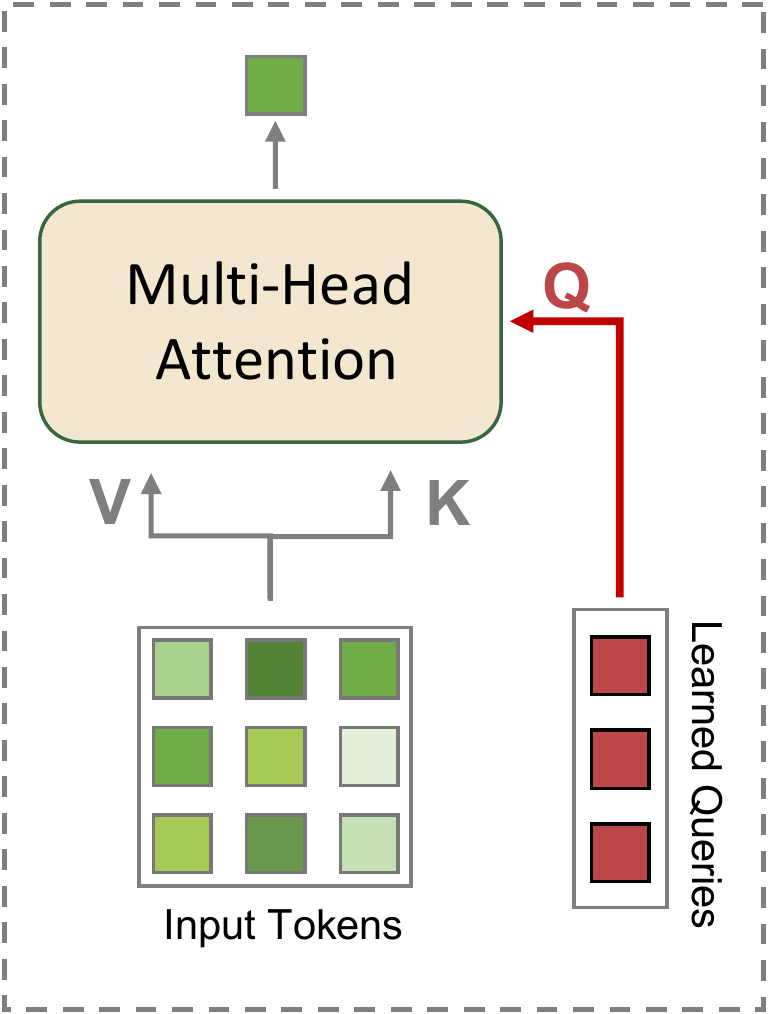}
        \subcaption{QnA (\textbf{ours})}
        \label{subfig:overview_ours}
    \end{subfigure}
    \caption{\textbf{QnA Overview.} Local layers operate on images by considering  overlapping windows (left), where the output is computed by aggregating information within each window: (a) Convolutions apply aggregation by learning weighted filters that are applied on each window. (b) Stand-Alone-Self-Attention (SASA) combines the window tokens via self-attention~\cite{SASA} --- 
    a time and memory consuming operation. (c) Instead of attending all window elements with each other,  
    we employ learned queries that are shared across windows. This allows linear space complexity, while maintaining the expressive power of the attention mechanism.}
    \label{fig:overview}
\end{figure*}

\section{Method}
\label{sec:method}
Query and Attend is a context-aware local feature processing layer. The key design choice of QnA is a convolution-like operation in which aggregation kernels vary according to the context of the processed local region. The heart of QnA is the attention mechanism, where overlapping windows are efficiently processed to maintain shift-invariance. Recall that three primary entities are deduced from the input features in self-attention: queries, keys, and values. The query-key dot product, which defines the attention weights, can be computationally pricey. To overcome this limitation, we detour from extracting the queries from the window itself but learn them instead (see~\Cref{subfig:overview_ours}).  This process is conceptually similar to convolution kernels, as the learned queries determine how to aggregate token values, focusing on feature subspaces pre-defined by the network. We show that learning the queries maintains the expressive power of the self-attention mechanism and facilitates a novel efficient QnA implementation that uses only simple and fast operations.  Finally, our layer can be extended to perform other functionalities (e.g., downsampling and upsampling), which are non-trivial in existing methods~\cite{SASA, HaloNet}.

Before the detailed explanation of QnA, we will briefly discuss the benefits and limitations of convolutions and self-attention. We let $H$ and $W$ be the height and width of the input feature maps, and denote $D$ as the embedding dimension. Otherwise, throughout this section, we use upper-case notation to denote a matrix or tensor entities, and lower-case notation to denote scalars or vectors.

\subsection{Convolution}
The convolution layer aggregates information by considering a local neighborhood of each element (e.g., a pixel) of the input feature $X \in \mathbb{R}^{H\times W \times D}$. Specifically, given a kernel $W \in \mathbb{R}^{k\times k \times D \times D}$, the convolution output at location $(i,j)$ is:
\begin{equation}
\label{eq:convolution}
    z_{i,j} = \sum_{\substack{(n,m) \in \\ \mathcal{N}_{k}(i,j)}} x_{n,m}\cdot W_{\lfloor k/2 \rfloor+i-n,\lfloor k/2 \rfloor+j-m},
\end{equation}
where the $k\times k$-spatial neighborhood of location $(i,j)$ is $$\mathcal{N}_{k}(i,j) = \{ (n,m) | -k/2 < (i-n), (j-m) \leq k/2  \}$$ (see~\Cref{subfig:overview_conv}). To simplify the notation, we omit $k$ from~\Cref{eq:convolution} and re-write it in matrix notation as:
\begin{equation}
\label{eq:convolution_simple}
    z_{i,j} = X_{\mathcal{N}_{i,j}} \cdot W,
\end{equation}
For brevity, we assume a stride 1 for all strided operations, and padding is applied to maintain spatial consistency.

The number of convolutional parameters is quadratic in kernel size, inhibiting usage of large kernels, therefore limiting the ability to capture global interactions. In addition, reusing convolutional filters across different locations does not allow adaptive content-based filtering. Nevertheless, the locality and shift-invariance properties of convolutions benefit vision tasks. For this reason, convolutions are widely adopted in computer vision networks, and deep learning frameworks support hardware-accelerated implementation of~\Cref{eq:convolution}.

\subsection{Self-Attention}
A vision transformer network processes a sequence of $D$-dimensional vectors, $X \in \mathbb{R}^{N\times D}$, by mixing the sequence of size $N$ through the self-attention mechanism. These vectors usually encode some form of image patches where $N = H\times W$ and $H,W$ are the number of patches in each spatial dimension. Specifically, the input vectors are first projected into keys $K=X W_K$, values $V=X W_V$ and queries $Q=X W_Q$ via three linear projection matrices $W_K,  W_V, W_Q \in \mathbb{R}^{D\times D}$. Then, the output of the self-attention operation is defined by:
\begin{equation}
\label{eq:self-attention}
    \begin{split}
    \text{\textbf{SA}}(X) & = \text{\textbf{Attention}}\left(Q,K\right) \cdot V \\ 
          & = \text{\textbf{Softmax}}\left(Q K^{T} / {\sqrt{D}} \right) \cdot V,    
    \end{split}
\end{equation} where $\text{\textbf{Attention}}\left(Q,K\right)$ is an attention score matrix of size $N \times N$ which is calculated using Softmax that is applied over each row.

Unlike convolutions, self-attention layers have a global receptive field and can process the whole input sequence, without affecting the number of learned parameters. Furthermore, every output of the self-attention layer is an input-dependent linear combination of the $V$ values, whereas in convolutions the aggregation is the same across the spatial dimension. However, the self-attention layer suffers from quadratic run-time complexity and inefficient memory usage, which makes it less favorable for processing high-resolution inputs. Furthermore, it has been shown that vanilla transformers don't attend locally very well ~\cite{ViT, DEiT, ConViT, CNNvsVIT}, which is a desired prior for downstream tasks. These models tend to become more local in nature only after a long and data-hungry training process~\cite{CNNvsVIT}. 

\subsection{Query-and-Attend}
To devise a high-powered layer, we will adapt the self-attention mechanism into a convolution-like aggregation operation. The motivation behind this is that, as it has already been shown~\cite{CoAtNet, ViT}, self-attention layer has better capacity than the convolution layer, yet, the inductive bias of convolutions allows better transferability and generalization capability~\cite{CoAtNet}. Specifically, the locality and shift-invariance priors (for early stages) impose powerful guidance in the image domain.

We begin by revisiting the Stand-Alone-Self-Attention approach (SASA)~\cite{SASA}, where attention is computed in small overlapping $k\times k$-windows, much like a convolution. The output $z_{i,j}$ of SASA is defined as:

\begin{equation}
\label{eq:sasa}
    z_{i,j}  = \textbf{Attention}\left(q_{i,j},K_{\mathcal{N}_{i,j}}\right) \cdot V_{\mathcal{N}_{i,j}},
\end{equation} where $q_{i,j} = X_{i,j} W_Q$. In other words, in order to aggregate tokens locally, self-attention is applied between the tokens of each local window, and a single query is extracted from the window center (see~\Cref{subfig:overview_sasa}).

While SASA~\cite{SASA} enjoy expressiveness and locality, through an input-adaptive convolution-like operation, it demands heavy memory usage. Specifically, to the best of our knowledge, all publicly available implementations use an unfolding operation that extracts patches from the input tensor. This operation expands the memory requirement by $k^2$ if implemented naively. Vaswani et al.~\cite{HaloNet} improved the memory-requirement of SASA~\cite{SASA} using local attention with halo expansion. Nevertheless, this implementation requires x3-x10 more memory than QnA while being x5-x8 slower, depending on $k$ (see~\Cref{fig:complexity_comparison}). This limitation makes the SASA layer infeasible for processing high-resolution images, employing larger kernels, or using sizable batches.

\subsubsection{QnA - Single Query}
To alleviate the compute limitation of SASA~\cite{SASA}, we redefine the key-query dot product in~\Cref{eq:sasa} by introducing learned queries. As we will later see, this modification leverages the weight-sharing principle (just like convolutions) and enables the efficient implementation of the QnA layer (see~\Cref{sec:implementation}).

We begin by first replacing the queries $q_{i,j}$ from~\Cref{eq:sasa} with a single $D$-dimensional vector $\tilde{q}$, that is learned during training. More particularly, we define the output of the QnA layer at location $(i,j)$ to be:
\begin{equation}
\label{eq:single_qna}
    z_{i,j}  = \textbf{Attention}\left(\tilde{q},K_{\mathcal{N}_{i,j}}\right) \cdot V_{\mathcal{N}_{i,j}}.
\end{equation}

Through the above modification, we interpret the query-key dot product as the scalar-projection of the keys onto D-dimensional query directions. Therefore, the token values are aggregated according to their relative orientation with the query vectors.  Intuitively, the keys can now be extracted such that relevant features' keys will be closely aligned with $\tilde{q}$. This means that the network can optimize the query direction to detect contextually related features.

\subsubsection{QnA - Multiple Queries}
As it turns out, performance can be further pushed forward under our paradigm, with minimal computational overhead and negligible additional memory. The naive approach is to add channels or attention heads when considering multi-head attention. While this enhances expressiveness, additional heads induce a larger memory footprint and computational overhead. To improve the layer expressiveness, we can use $L$-different queries $\tilde{Q} \in \mathbb{R}^{L\times D}$ instead of one. Nevertheless, simply plugging in $\tilde{Q}$ in~\Cref{eq:single_qna} leads to $L\times D$ output, which expands the memory usage by $L$ (also known as cross-attention). Instead, we weight-sum the attention maps learned by the queries into a single attention map (for each attention head) and use it to aggregate the values. Therefore our QnA output becomes:
\begin{equation}
\label{eq:mutliple_qna}
    z_{i,j}  = \left(\sum_{i \in \left[L\right]} \mathcal{W}_{i}*\textbf{Attention}\left(\tilde{Q}_i,K_{\mathcal{N}_{i,j}}\right) \right) \cdot V_{\mathcal{N}_{i,j}},
\end{equation} where $\mathcal{W} \in \mathbb{R}^{L \times k^2}$ is a learned weight matrix, and $*$ is the element-wise multiplication operation. The overall extra space used in this case is $\mathcal{O}(L \times k^2)$, which is relatively small, as opposed to the naive solution, which requires $\mathcal{O}(L\times D)$ extra space.

\subsubsection{QnA Variants}

Our layer naturally accommodates the improvements made for the vanilla self-attention layer~\cite{AttentionIsAlYouNeed}. Specifically, we use relative-positional embedding~\cite{RPE1, RPE2, RPE3, RPE4, RethinkingPE} and multi-head attention in all our models (further details can be found in~\Cref{sup:qna_variants}).

\paragraph{Upsampling \& Downsampling Using QnA} down-sampling can be trivially attained using strided windows. To up-scale tokens by a factor $s$, we can use a QnA layer with $L=s^2$ learned queries. Assigning the result of each query as an entry in the upsampled output, we effectively construct a spatially dynamic upsampling kernel of size $s\times s$. We define the upsampling operation more formally in~\Cref{sup:qna_variants}. We show that QnA could be used to efficiently perform the upsampling function (\Cref{sec:qna_upsample}) with improved performance, suggesting it can be incorporated into other vision tasks such as image synthesis.

\begin{figure}
    \centering
    \begin{subfigure}{0.45\linewidth}
        \centering
        \includegraphics[width=0.95\textwidth]{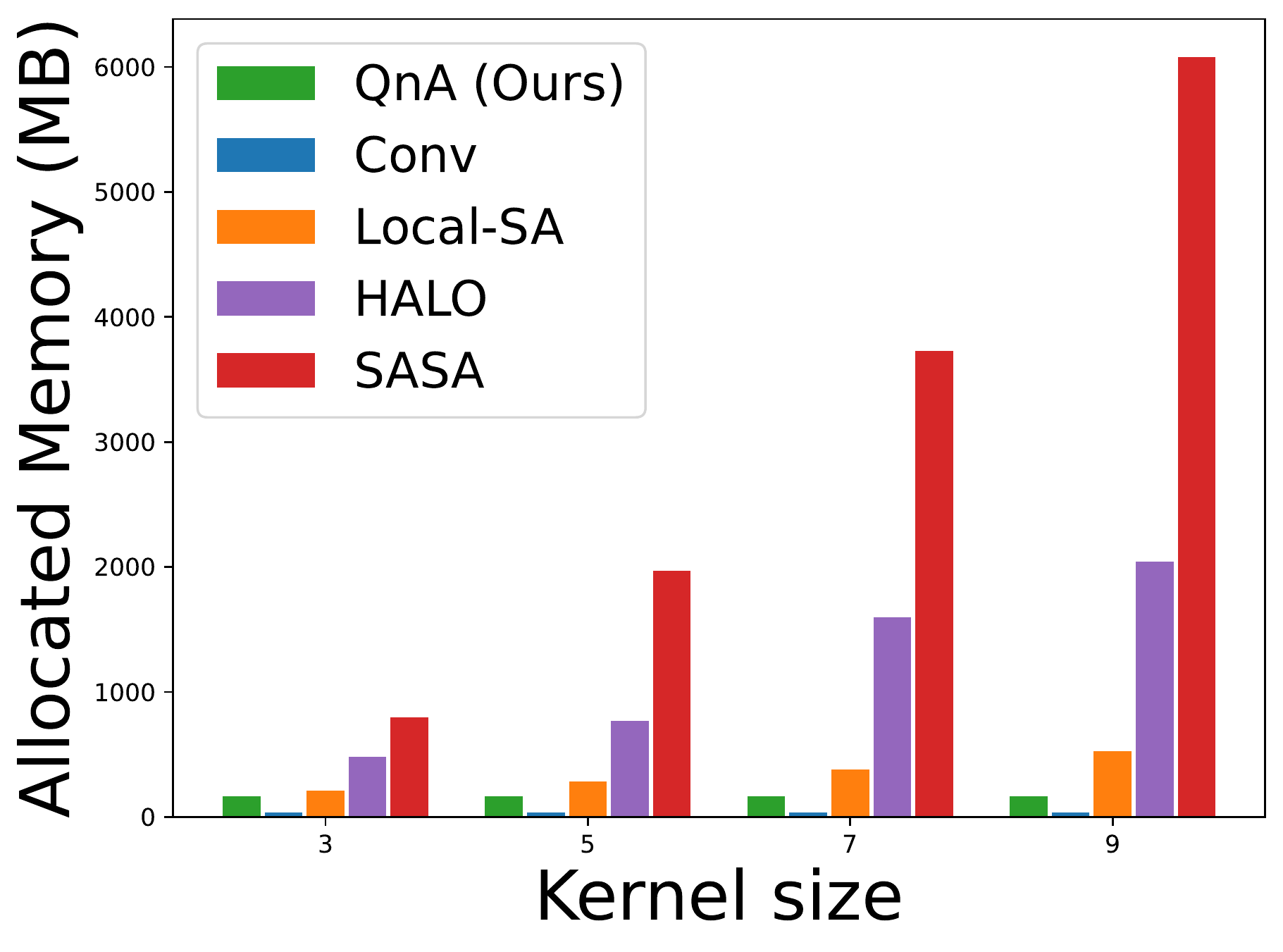}
        \caption{Memory allocation}
    \end{subfigure}
    \begin{subfigure}{0.45\linewidth}
        \centering
        \includegraphics[width=0.95\textwidth]{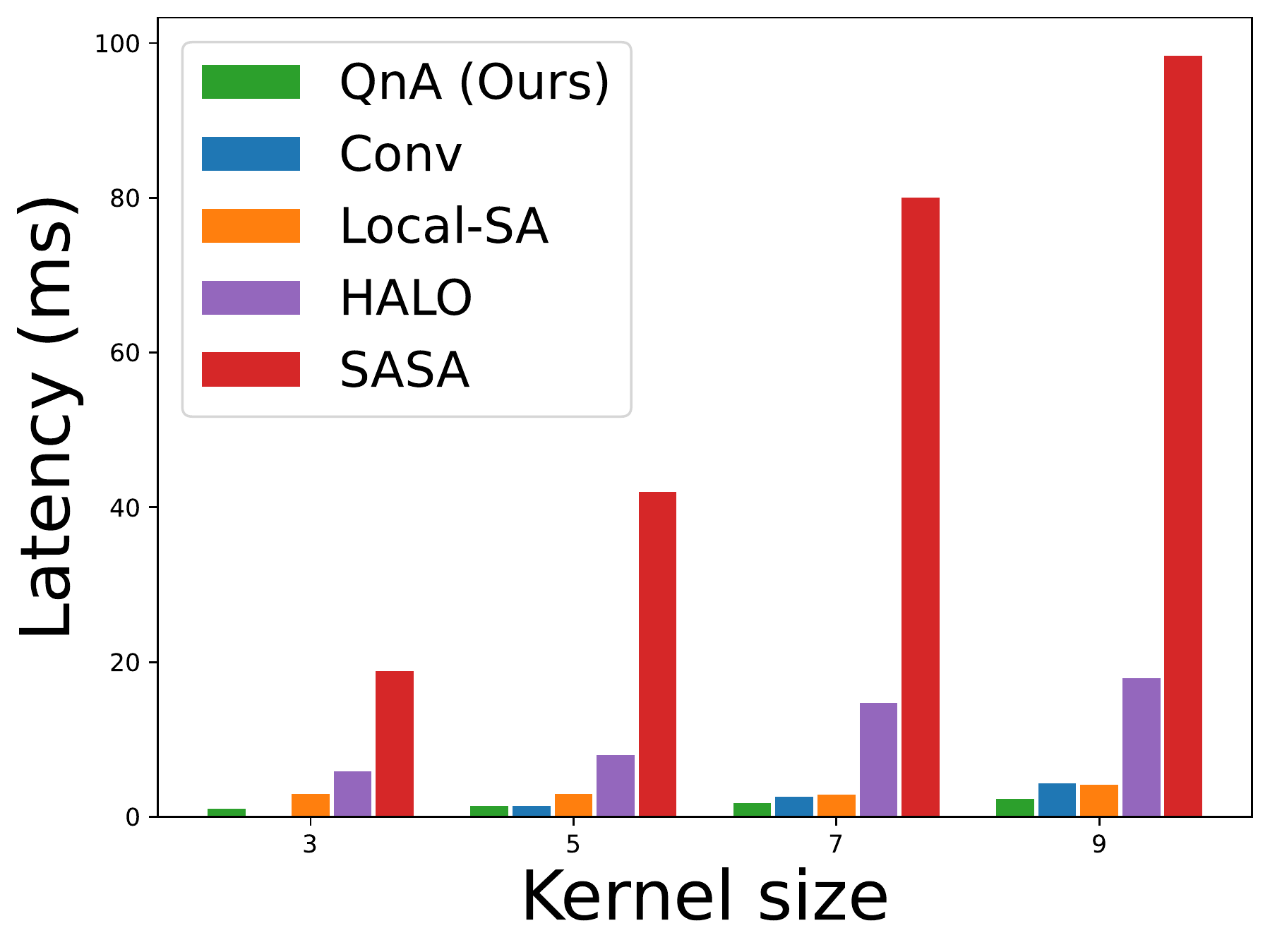}
        \caption{Latency}
    \end{subfigure}
    \caption{\textbf{Single Layer Computational Complexity During Forward Pass.} QnA outperforms SASA~\cite{SASA}, HaloNet~\cite{HaloNet}, and local self-attention baselines in terms of speed and memory consumption. In particular, during forward pass, HaloNet~\cite{HaloNet} requires at least x3 additional memory allocation while being x5 slower. For larger kernels, the computation overhead becomes significant where up-to x10 additional memory allocation is needed. Convolutional layers are the most memory efficient, however they are x1.8 slower compared to QnA for large kernels. All experiments tested with PyTorch~\cite{PyTorch}, on input size $256\times 256 \times 64$.}
    \label{fig:complexity_comparison}
\end{figure}
\subsection{Implementation \& Complexity Analysis}
\label{sec:implementation}
The shared-learned queries across windows allow us to implement QnA using efficient operations that are available in existing deep-learning frameworks (e.g., Jax~\cite{jax2018github}). In particular, the query-key dot product can be calculated once on the whole input sequence, avoiding extra space allocation. Then, we can use window-based operations to effectively calculate the softmax operation over the overlapping windows, leading to a linear time-and-space complexity (see~\Cref{fig:complexity_comparison}). Full-implementation details of our method are in~\cref{sup:implementation}, along with a code snippet in Jax/Flax~\cite{jax2018github, flax2020github}. 

\subsection{The QnA-ViT Architecture}
The QnA-ViT architecture is composed of vision transformer blocks~\cite{ViT} (for global context) and QnA blocks (for local context). The QnA block shares a similar structure with the ViT block, except we replace the multi-head self-attention layer with the QnA layer. We present a family of architectures that follow the design of ResNet~\cite{ResNet}. Specifically, we use a 4-stage hierarchical architecture. The base dimension $D$ varies according to the model size. Below we indicate how many layers we use in each stage ($T$ stands for ViT-blocks and $Q$ stands for QnA-blocks):
\begin{itemize}
 \item Tiny: $D,T,Q\!=\{64, \left[0,0,4,2\right],  \left[3,4,3,0\right] \}$
 \item Small: $D,T,Q\!=\{64, \left[0,0,12,2\right],  \left[3,4,7,0\right] \}$
 \item Base: $D,T,Q\!=\{96, \left[0,0,12,2\right],  \left[3,4,7,0\right] \}$
\end{itemize}
For further details, please refer to~\cref{sup:arch}.
\section{Experiments}

\subsection{Image Recognition \& ImageNet-1K Results}
\begin{table}
\centering
\resizebox{0.9\columnwidth}{!}{%
\begin{tabular}{@{}>{\columncolor{white}[0pt][\tabcolsep]}l|ccc|>{\columncolor{white}[\tabcolsep][0pt]}c@{}}
\addlinespace
\toprule[1.5pt]
\multicolumn{1}{r|}{Method}  & Params & GFLOPS & Throughput & Top-1 Acc. \\ \midrule
ResNet50~\cite{ResNet, ResNetStrikeBack} & 26M & 4.1 & \textbf{1287} & 80.4 \\
ResNet101~\cite{ResNet, ResNetStrikeBack} & 45M & 7.9 &  770 & 81.5 \\
ResNet152~\cite{ResNet, ResNetStrikeBack} & 60M & 11.6 &  539 & 82.0 \\ \midrule
DeiT-S~\cite{DEiT}           & 22M    & 4.6    & 940 & 79.8           \\
DeiT-B~\cite{DEiT}           & 86M    & 17.5   & 292 & 81.8           \\ \midrule
Swin-Tiny~\cite{Swin}                & 29M    & 4.5    & 723 & 81.3           \\
Swin-Small~\cite{Swin}               & 50M    & 8.7    & 425 & 83.0           \\
Swin-Base~\cite{Swin}                & 88M    & 15.4   & 277 & 83.5           \\
Swin-Base~\cite{Swin}${\scriptstyle \uparrow 384}$  & 88M    & 47.0   & 85 & 84.5           \\
\midrule
NestT-Tiny~\cite{NestT}        & 17M    & 5.8    & 568 & 81.5           \\
NestT-Small~\cite{NestT}       & 38M    & 10.4    & 352 & 83.3           \\
NestT-Base~\cite{NestT}        & 68M    & 17.9   & 233 & 83.8           \\ \midrule
Focal-Tiny~\cite{Swin}            & 29M    & 4.9    & 546 & 82.2           \\
Focal-Small~\cite{Swin}           & 51M    & 9.1    & 282 & 83.5           \\
Focal-Base~\cite{Swin}            & 90M    & 16.0   & 207 & 83.8           \\ \midrule
\rowcolor{lightgreen} QnA-Tiny              & \textbf{16M}    & 2.5    & 1060 & 81.7           \\
\rowcolor{lightgreen} QnA-Tiny${\scriptstyle 7\times 7}$              & 16M    & 2.6    & 895 & 82.0           \\
\rowcolor{lightgreen} QnA-Small             & 25M    & 4.4    & 596 & 83.2           \\
\rowcolor{lightgreen} QnA-Base              & 56M    & 9.7    & 372 & 83.7           \\
\rowcolor{lightgreen} QnA-Base${\scriptstyle \uparrow 384}$    & 56M    & 30.6  & 177 & \textbf{84.8}           \\
\bottomrule[1.5pt]
\end{tabular}
}
\caption{\textbf{ImageNet-1K\cite{ImageNet} pre-training results.} All models were pre-trained and tested on input size $224\times 224$. Models marked with $\uparrow 384$ are later also fine-tuned and tested on $384^2$ resolution, following~\cite{FixResNet}. The Accuracy, parameter count, and floating point operations are as reported in the corresponding publication. Throughput was calculated using the timm~\cite{timm} library, on a single NVIDIA V100 GPU with 16GB memory. For QnA${\scriptstyle 7\times 7}$, a $7\times 7$ window size was used instead of $3 \times 3$. Our model achieves comparable results to state-of-the-art models, with fewer parameters and better computation complexity.}
\label{tbl:main_classification_results}
\end{table}
\paragraph{Setting:} we evaluate our method using the ImageNet-1K~\cite{ImageNet} benchmark, and follow the training recipe of DEiT~\cite{DEiT}, except we omit EMA~\cite{EMA} and repeated augmentations~\cite{RepeatedAug}. For full-training details please refer to~\cref{sup:training_details}. 

\paragraph{Results:} A summary comparison between different models appears in~\Cref{tbl:main_classification_results}. As shown from the table, most transformer-based vision models outperform CNN-based ones in terms of the top-1 accuracy, even when the CNN models are trained using a strong training procedure. For example, ResNet50~\cite{ResNet} with standard ImageNet training achieves 76.6\% top-1 accuracy. However, as argued in~\cite{ResNetStrikeBack}, with better training, its accuracy sky-rockets to 80.4\%. Indeed, this is a very impressive improvement, yet it falls short behind transformer models. In particular, our model (the tiny version) improves upon ResNet by 1.3\% with ~40\% fewer parameters and FLOPs. 

In terms of speed, CNNs are very fast and have a smaller memory footprint (see~\Cref{fig:complexity_comparison}). The throughput gap can be evident by investigating the vision transformers reported in Table~\ref{tbl:main_classification_results}. A particular strong ViT is the Focal-ViT~\cite{FocalViT}; in its tiny version, it improves upon ResNet101 by 0.7\% while the latter enjoys x1.4-times better throughput. Nonetheless, our model stands out in terms of the speed-accuracy trade-off. Comparing QnA-Tiny with Focal-Tiny, we achieve only 0.5\% less accuracy while having x2-times better throughput, parameter-count, and flops. We can even reduce this gap by training the QnA with a larger receptive field. For example, setting the receptive field of the QnA to be 7x7, instead of 3x3, achieve 82.0\% accuracy, with negligible effect on the model speed and size. 

Finally, we notice that most Vision Transformers achieve similar Top-1 accuracy. More specifically, tiny models (in terms of parameters and number of FLOPs) achieve roughly the same Top-1 accuracy of 81.2-82.0\%. The accuracy difference is even less significant in larger models (e.g., base variants accuracy differs by only 0.1\%), and this accuracy difference can be easily tipped to either side by many factors, even by choosing a different seed~\cite{Seed}. Nonetheless, our model is faster, all while using fewer resources. 

\paragraph{The reason behind better accuracy-efficiency trade-off:} QnA-ViT achieves a better accuracy-efficiency trade-off for several reasons. First, QnA is fast, which is crucial for better throughput. Further, most of the vision transformer's parameter count is due to the linear projection matrices. Our method reduces the number of linear projections by omitting the query projections (i.e., the $W_q$ matrix is replaced with 2-learned queries). Furthermore, the feed-forward network requires $\times2$ more parameters than the self-attention. Our model uses smaller embedding dimensions than existing models without sacrificing accuracy. Namely, NesT-Tiny~\cite{NestT} uses an embedding dimension of 192, while Swin-Tiny~\cite{Swin} and Focal-Tiny~\cite{FocalViT} use 96 embedding dimensions. On the other hand, our method achieves a similar feature representation capacity, with a lower dimension of 64.

Finally, other parameter efficient methods achieve low parameter count by training on larger input images~\cite{HaloNet, EfficientNet}. This is shown to improve image-classification accuracy~\cite{FixResNet}. However, it comes at the cost of lower-throughput and more FLOPs. For example, EfficientNet-B5~\cite{EfficientNet}, which was trained and tested on images of $456\times 456$ resolution, achieves 83.6\% accuracy while using only 30M parameters. Nonetheless, the network's throughput is 170 images/sec, and it uses 9.9 GFLOPs. Compared to our base model, QnA achieves similar accuracy with twice the throughput. Also, it is important to note that these models were optimized via Neural Architecture search, an automated method for better architecture design. We believe employing methods with similar purpose~\cite{NViT} would even further optimize our models' parameter count.

\begin{table}
\centering
\resizebox{0.7\linewidth}{!}{%
\begin{tabular}{@{}>{\columncolor{white}[0pt][\tabcolsep]}lc
>{\columncolor{lightgreen}}c 
>{\columncolor{lightgreen}}c 
>{\columncolor{lightgreen}}c 
>{\columncolor{lightgreen}}c >{\columncolor{white}[0pt][\tabcolsep]}c@{}}
\toprule[1.5pt]
                        &       & \multicolumn{4}{c}{\cellcolor{lightgreen}QnA}     \\
                        & SASA  & $L=1$          & $L=2$    & $L=3$       & $L=4$       \\ \midrule[1.5pt]
Top-1 Acc.              & \textbf{80.86} & 80.3         & 80.7       & 80.76     & 80.81     \\ \midrule
Params (M).             & 16.440& \textbf{16.182}       & 16.188     & 16.192   & 16.200.   \\ \midrule
FLOPS (G)               & 2.620 & \textbf{2.378}        & 2.400      & 2.420     & 2.442     \\ \bottomrule[1.5pt]
\end{tabular}%
}
\caption{\textbf{Multiple queries effect.} We compare the performance of SASA~\cite{SASA} to QnA with a varying amount of queries. As can be seen,  using multiple queries improves QnA, reaching comparable performance, using an order of magnitude less memory.}
\label{tbl:sasa_multiple_queries}
\end{table}

\subsection{Ablation \& Design Choices}
\label{sec:ablation_and_design_choices}

\paragraph{Number of Queries:} Using multiple queries allows us to capture different feature subspaces. We consider SASA~\cite{SASA} as our baseline, which extracts the self-attention queries from the window elements. Due to its heavy memory footprint, we cannot consider SASA variants similar to QnA-ViT. Instead, we consider a lightweight variant that combines local self-attention with SASA. All SASA layers use a 3x3 window size. Downsampling is performed similar to QnA-ViT, except that we replace QnA with SASA. Finally, the local-self attention layers use a 7x7 window size without overlapping (see~\Cref{sup:num_of_queries}). The results are summarized in Table~\ref{tbl:sasa_multiple_queries}. As can be seen, we achieved comparable results to SASA. In addition, two queries outperform one, but this improvement saturates quickly. We hence recommend using two queries, as it enjoys efficiency and expressiveness. 

\paragraph{Number of heads:} Most vision transformers use large head dimension (e.g., $\geq 32$)\cite{CaiT}. However, we found that the QnA layer enjoys more heads. We trained various models based on QnA and self-attention layers with different training setups to verify this. Our experiments found that a head dimension $d=8$ works best for QnA layers. Similar to previous work~\cite{CaiT}, in hybrid models, where both self-attention and QnA layers are used, we found that self-attention layers still require a large head dimension (i.e., $d=32$). Moreover, we found that using more heads for QnA is considerably better (up to $1 \%$ improvement) for small networks. Moreover, this performance gap is more apparent when training the models for fewer epochs without strong augmentations (see~\Cref{sup:num_of_heads} for further details). Intuitively, since the QnA layer is local, it benefits more from local pattern identifications, unlike global context, which requires expressive representation. 

\begin{table}
\resizebox{\linewidth}{!}{%
\begin{tabular}{@{}>{\columncolor{white}[0pt][\tabcolsep]}ccccc|>{\columncolor{white}[\tabcolsep][0pt]}c@{}}
\toprule[1.5pt]
Global Attention              & QnA              & Downsampling & Params & FLOPs & Top1-Acc. \\ \midrule[1.1pt]
\multicolumn{6}{c}{Different downsampling choices}                              \\ \midrule[1.1pt]
{[}3,3,6,2{]}  & {[}0,0,0,0{]}  & Nest~\cite{NestT}         & 16.8M  & 3.7   & 81.2          \\
{[}3,3,6,2{]}  & {[}0,0,0,0{]}  & Swin~\cite{Swin}         & \textbf{16.0M}  & \textbf{3.1}   & 81.2          \\
{[}3,3,6,2{]}  & {[}1,1,1,0{]}  & QnA          & \textbf{16.0M} & 3.2   & \textbf{81.9}          \\ \midrule[1.1pt]
\multicolumn{6}{c}{Number of QnA blocks vs Transformer blocks}                                                \\ \midrule[1.1pt]
{[}0,0,0,0{]}  & {[}4,4,7,2{]}  & QnA          & 14.9M  & 2.4   & 80.9          \\
{[}3,3,6,2{]}  & {[}1,1,1,0{]}  & QnA          & 16.0M  & 3.2   & \textbf{81.9}          \\
\rowcolor{lightgreen} 
{[}0,0,4,2{]}  & {[}4,4,3,0{]}  & QnA          & \textbf{15.8M}  & \textbf{2.6}   & \textbf{81.9}          \\ \midrule[1.1pt]
\multicolumn{6}{c}{Deeper Models}                                               \\ \midrule[1.1pt]
{[}0,0,8,2{]}  & {[}3,4,11,0{]} & QnA          & \textbf{24.7M}      & \textbf{4.2}     & 82.7          \\
{[}0,0,10,2{]} & {[}3,4,9,0{]}  & QnA          & 24.8M      & 4.3     & 83.0          \\
\rowcolor{lightgreen} 
{[}0,0,12,2{]} & {[}3,4,7,0{]}  & QnA          & 25.0M      & 4.4     & \textbf{83.2}          \\
{[}0,0,16,2{]} & {[}3,4,3,0{]}  & QnA          & 25.3M      & 4.6     & 83.1          \\ \bottomrule[1.5pt]
\end{tabular}%
}
\caption{\textbf{Ablation studies and design choices.} In the first two columns we specify the number of global-attention and QnA layers used in each stage. See~\Cref{sec:ablation_and_design_choices} for further details, and the supp. materials for more configurations.}
\label{tbl:design_choices}
\end{table}
\paragraph{How many QnA layers do you need?} 
In order to verify the expressive power of QnA, we consider a dozen different models. Each model consists of four stages. In each stage, we consider using self-attention and QnA layers. A summary report can be found in Table~\ref{tbl:design_choices} (for the full report, please see~\Cref{sup:how_much_qna}). In our experiments, we conclude that the QnA layer is effective in the early stages and can replace global attention without affecting the model's performance. QnA is fast and improves the model's efficiency. Finally, the QnA layer is a very effective down-sampling layer. For example, we considered two baseline architectures which are mostly composed of transformer blocks, (1) one model uses simple 2x2 strided-convolution to reduce the feature maps (adopted in~\cite{Swin}), and the (2) other is based on the down-sampling used in NesT~\cite{NestT}, which is a 3x3 convolution, followed by a layer-norm and max-pooling layer. These two models achieve similar accuracy, which is 81.2\%. On the other hand, when merely replacing the downsampling layers with the QnA layer, we witness a ~0.7\% improvement without increasing the parameter count and FLOPs. Note, global self-attention is still needed to achieve good performance. However, it can be diminished by local operations, e.g., QnA.

\paragraph{Deep models:} 
To scale-up our model, we chose to increase the number of layers in the network's third stage (as typical in previous works~\cite{ResNet}). This design choice is adapted mainly for efficiency reasons, where the spatial and feature dimension are manageable in the third stage. In particular, we increase the total number of layers in the third stage from 7 to 19 and consider four configurations where each configuration varies by the number of QnA layers used. The models' accuracies are reported in Table~\ref{tbl:design_choices}. As seen from the table, the model's accuracy can be maintained by reducing the number of global attention. This indicates that while self-attention can capture global information, it is beneficial to a certain degree, and local attention could be imposed by the architecture design for efficiency consideration. 

\subsection{Object Detection}
\begin{table}
\resizebox{\linewidth}{!}{%
\begin{tabular}{@{}>{\columncolor{white}[0pt][\tabcolsep]}llccccc>{\columncolor{white}[\tabcolsep][0pt]}c@{}}
\toprule
Model    & Backbone & $\text{AP}_\text{50}$ & $\text{AP}_\text{75}$ & $\text{AP}_\text{L}$  & $\text{AP}_\text{M}$  & $\text{AP}_\text{S}$  & AP   \\ \midrule
         & R50      & 55.4 & 36.6 & 53.2 & 38.0 & 15.1 & 35.3 \\ \cmidrule(l){2-8} 
DETR &
  \cellcolor{lightgreen}QnA-Ti &
  \cellcolor{lightgreen}58.9 &
  \cellcolor{lightgreen}38.6 &
  \cellcolor{lightgreen}56.8 &
  \cellcolor{lightgreen}40.6 &
  \cellcolor{lightgreen}16.0 &
  \cellcolor{lightgreen}37.5 \\ \cmidrule(l){2-8} 
 &
  \cellcolor{lightgreen}QnA-Ti7 &
  \cellcolor{lightgreen}\textbf{59.6} &
  \cellcolor{lightgreen}39.3 &
  \cellcolor{lightgreen}\textbf{57.6} &
  \cellcolor{lightgreen}41.2 &
  \cellcolor{lightgreen}16.0 &
  \cellcolor{lightgreen}37.9 \\ \midrule
\rowcolor{lightgreen} 
DETR-QnA & QnA-Ti   & \textbf{59.6} & \textbf{39.7} & 57.4 & \textbf{41.8} & \textbf{18.2} & \textbf{38.5} \\ \bottomrule
\end{tabular}%
}
\caption{\textbf{DETR~\cite{DETR} Based Object detection on the COCO Dataset~\cite{COCO}}. Incorporating QnA-ViT-Tiny with DETR substantially improves upon the ResNet50 backbone (by up to 3.2). QnA with receptive field 7x7 improves the average precision on large objects ($\text{AP}_\text{L}$), and incorporating QnA into the DETR network improves performance on smaller objects, indicating locality. }
\label{tbl:detr}
\end{table}

\paragraph{Setting:} To evaluate the representation quality of our pre-trained networks, we use the DETR~\cite{DETR} framework, which is a transformer-based end-to-end object detection framework. We use three backbones for our evaluations; ResNet50~\cite{ResNet}, and two variants of QnA-ViT, namely, QnA-ViT-Tiny, and QnA-ViT-Tiny-7x7, which uses a 7x7 receptive for all QnA layers (instead of 3x3). Complete training details are provided in the supplemental material.

\paragraph{Revisiting DETR transformer design: } 
DETR achieves comparable results to CNN-based frameworks~\cite{FastRCNN}. However, it achieves less favorable average precision when tested on smaller objects. The DETR model uses a vanilla transformer encoder to process the input features extracted from the backbone network. As argued earlier, global attention suffers from locality issues. To showcase the potential of incorporating QnA in existing transformer-based networks, we propose DETR-QnA architecture, in which two transformer blocks are replaced with four QnA blocks.

\paragraph{Results: } We report the results in Table~\ref{tbl:detr}. As can be seen, DETR trained with QnA-Tiny achieves +2.2 better AP compared to the ResNet50 backbone. Using a larger receptive field ($7\times7$) further improves the AP by 0.4. However, much improvement is due to better performance on large objects (+0.7). Finally, when incorporating QnA into the DETR encoder, we gain an additional +0.6AP (and +1.0AP relative to using the DETR model). More particularly, incorporating QnA with DETR achieves an impressive +2.2 AP improvement on small objects, indicating the benefits of QnA's locality.

\begin{figure}
    \centering
    \begin{subfigure}{0.3\linewidth}
        \centering
        \includegraphics[height=4.cm]{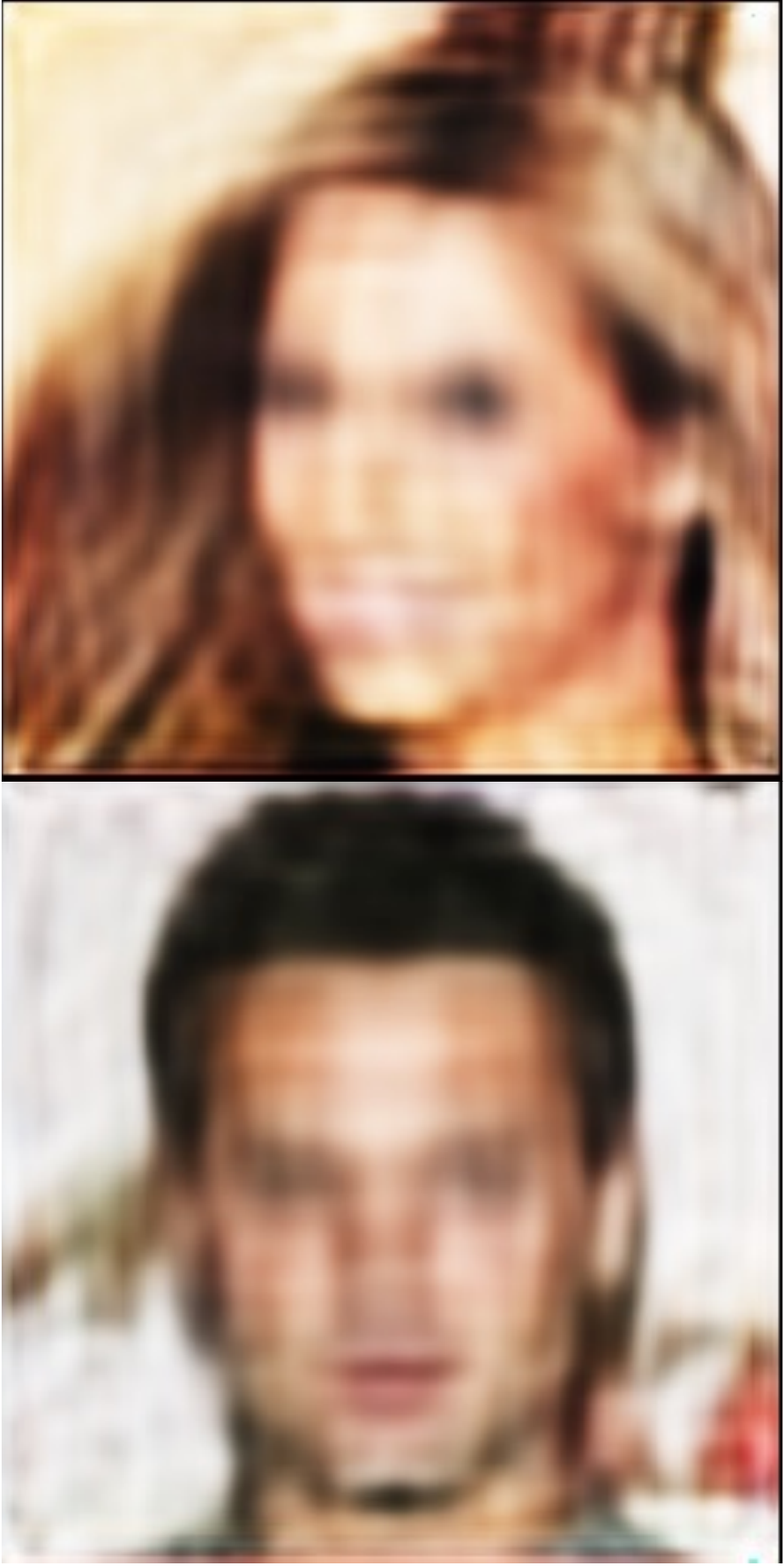}
        \caption{Bilinear}
    \end{subfigure}
        \begin{subfigure}{0.3\linewidth}
        \centering
        \includegraphics[height=4.cm]{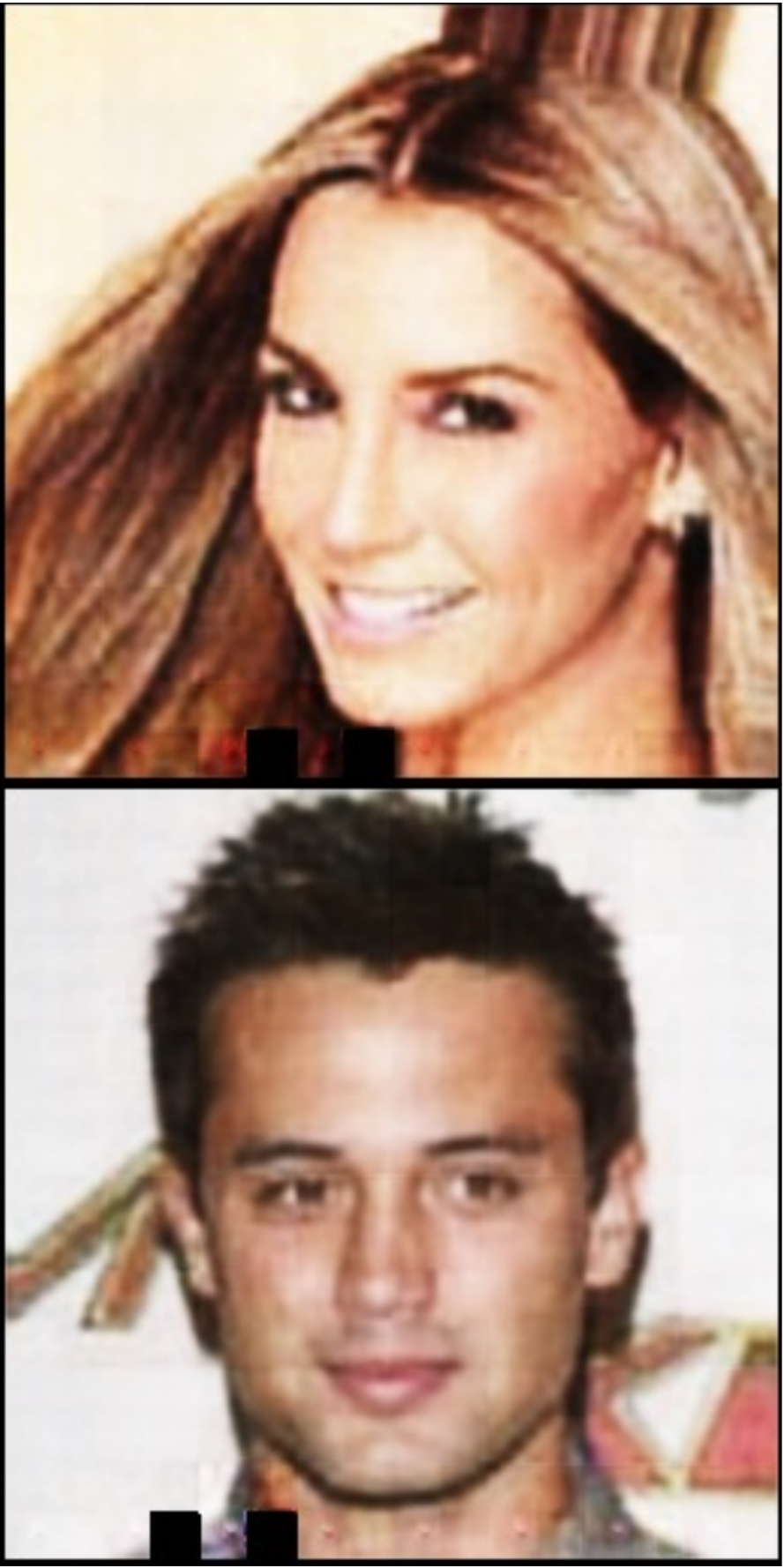}
        \caption{ConvTransposed}
    \end{subfigure}
        \begin{subfigure}{0.3\linewidth}
        \centering
        \includegraphics[height=4.cm]{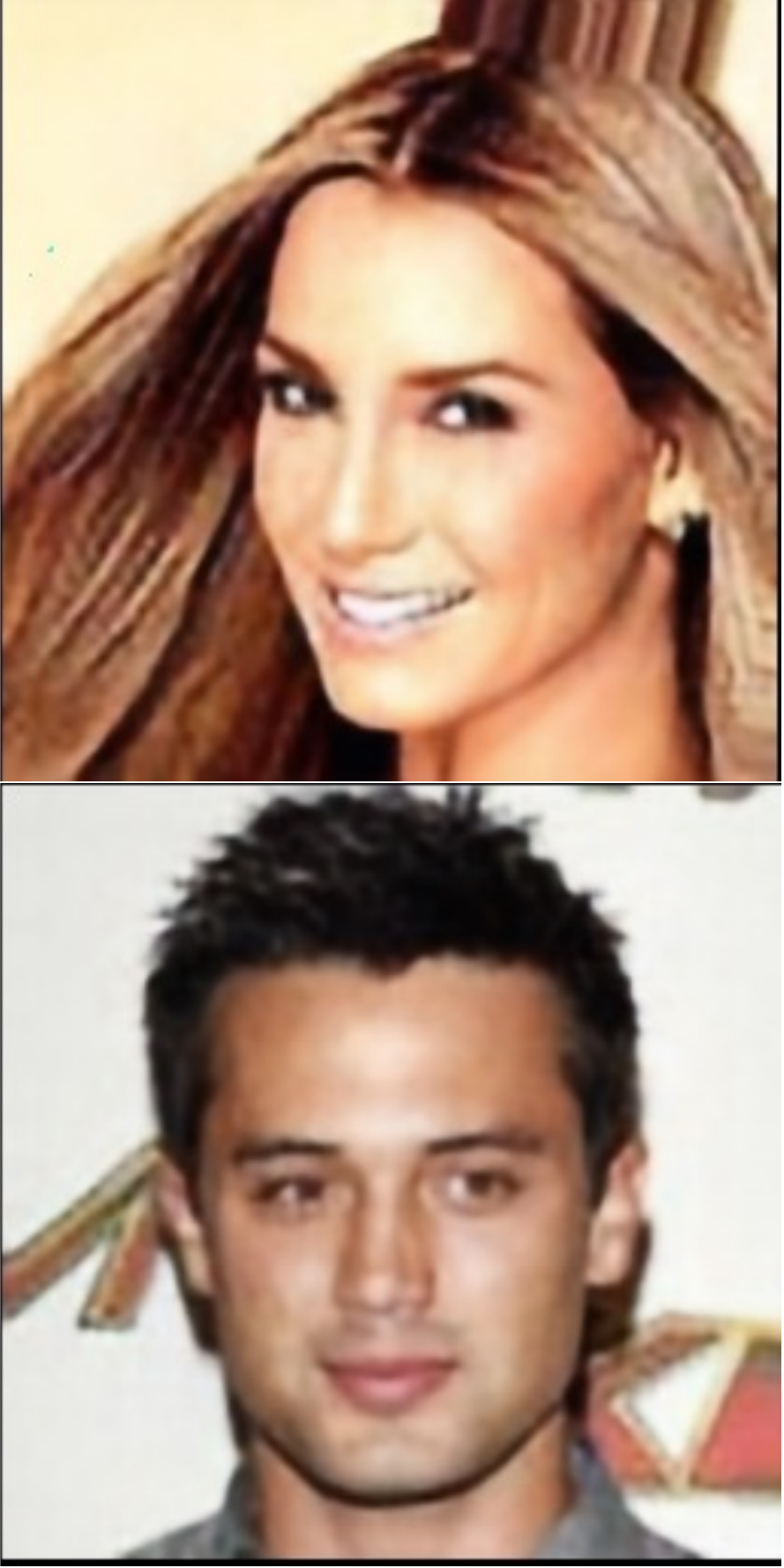}
        \caption{QnA}
    \end{subfigure}
    \caption{\textbf{Qualitative Auto-Encoder Results.} We train a simple Autoencoder using convolution layers (a-b), and (c) QnA layers. We show reconstructed images from the CelebA test set~\cite{CelebA}. QnA shows preferable reconstructions. See~\Cref{sec:qna_upsample} for more details.}
    \label{fig:upsample_qualitative}
\end{figure}
\subsection{QnA as an upsampling layer}
\label{sec:qna_upsample}
\begin{figure}
    \centering
    \begin{subfigure}{\linewidth}
        \centering
        \includegraphics[trim={0 1.4cm 0 0 }, clip, height=2.6cm]{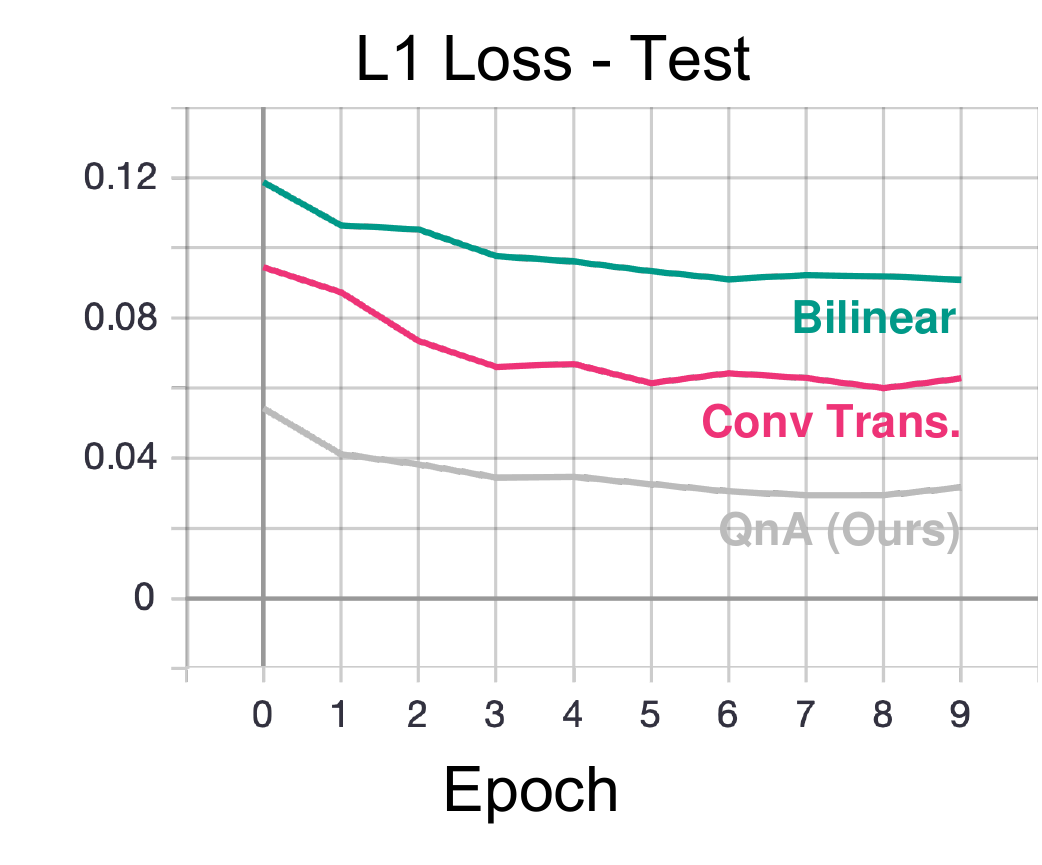}
        \includegraphics[trim={0 1.4cm 0 0 }, clip, height=2.6cm]{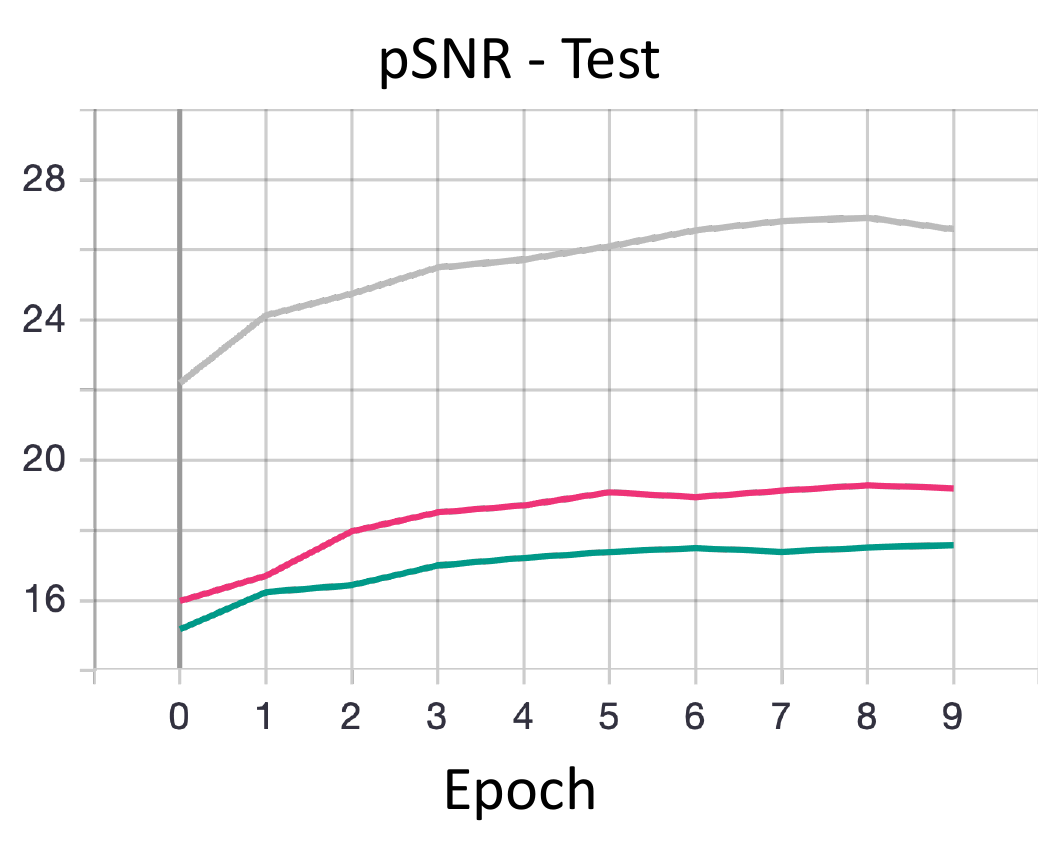}
    \end{subfigure}
        \begin{subfigure}{\linewidth}
        \centering
        \includegraphics[trim={0 1.4cm 0 0 }, clip, height=2.6cm]{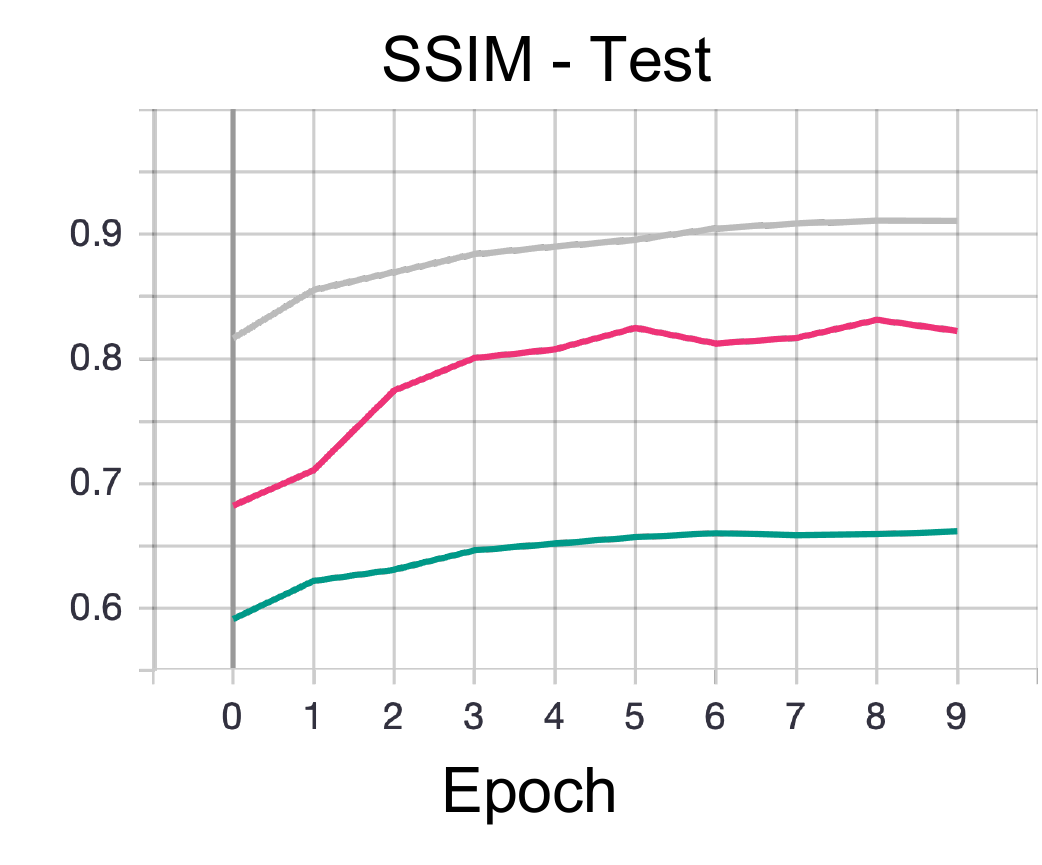}
        \includegraphics[trim={0 1.4cm 0 0 }, clip, height=2.6cm]{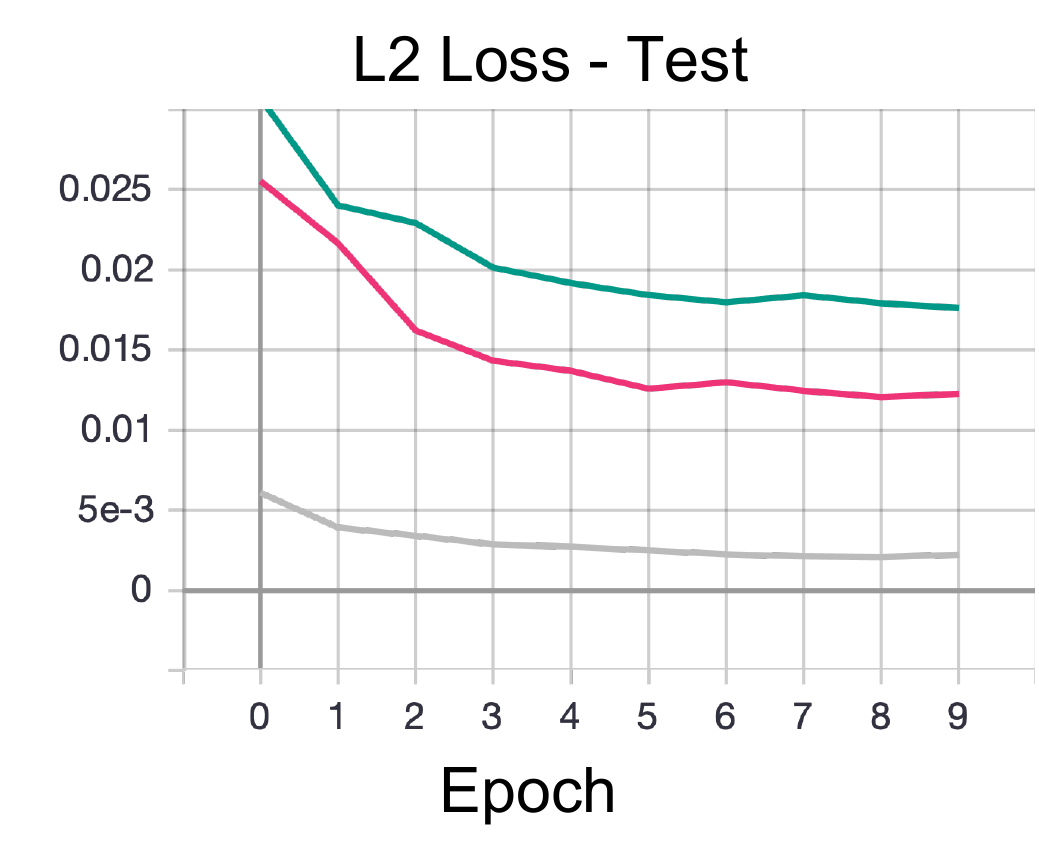}
    \end{subfigure}
    \caption{\textbf{Quantitative Auto-Encoder Results} are reported, compared to the same convolutional baselines as in Figure~\ref{fig:upsample_qualitative}. We compare our method (gray) to bilinear upsampling (green) and transposed covolution-based upsampling (pink). We show consistent improvement across epochs (horizontal axes) in the L1 loss (top left), the pSNR (top right), the SSIM (bottom left), and MSE (bottom right) metrics. 
    }
    \label{fig:upsample_quantitative}
\end{figure}

We suggest that QnA can be adapted to other tasks besides classification and detection. To demonstrate this, we train an autoencoder network on the CelebA~\cite{CelebA} dataset, using the $L_1$ reconstruction loss. We consider two simple baselines that are convolution-based. In particular, one baseline uses bilinear up-sampling to upscale the feature maps, and another baseline uses the transposed convolution layer~\cite{DeConv}. Qualitative and quantitative results appear in Figure~\ref{fig:upsample_qualitative} and Figure~\ref{fig:upsample_quantitative}, respectively. The figures show that the QnA-based auto-encoder achieves better qualitative and quantitative results and introduces fewer artifacts (see~\Cref{sup:auto-encoders} for further details). 
\section{Limitations \& Conclusion}
\label{sec:conclusion}

We have presented QnA, a novel local-attention layer with linear complexity that is also shift-invariant. Through rigorous experiments, we showed that QnA could serve as a general-purpose layer and improve the efficiency of vision transformers without compromising on the accuracy part. Furthermore, we evaluated our method in the object-detection setting and improved upon the existing self-attention-based method. Our layer could also be used as an up-sampling layer, which we believe is essential for incorporating transformers in other tasks, such as image generation. Finally, since QnA is attention-based, it requires additional intermediate memory, whereas convolutions operate seamlessly, requiring no additional allocation. Nonetheless, QnA has more expressive power than convolution. In addition, global self-attention blocks are more powerful in capturing global context. Therefore, we believe that our layer mitigates the gap between self-attention and convolutions and that future works should incorporate all three layers to achieve the best performance networks.

\section*{Acknowledgment}
Research supported with Cloud TPUs from Google's TPU Research Cloud (TRC).

{\small
\bibliographystyle{ieee_fullname}
\bibliography{egbib}
}

\appendix

\section{Attention Visualization}
\label{sup:attn_vis}
\begin{figure}
    \centering
    \includegraphics[width=.9\linewidth]{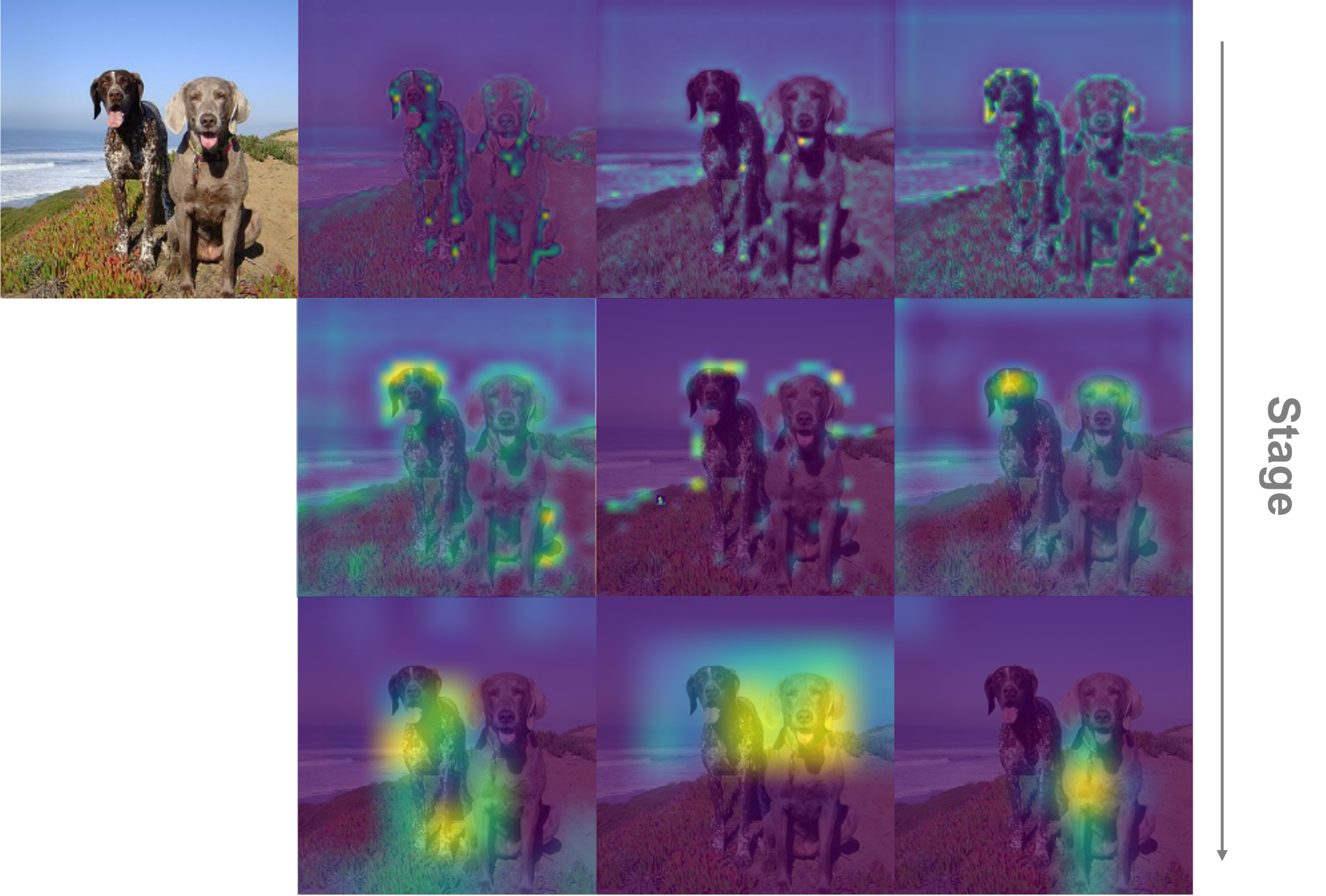}
    \caption{\textbf{QnA attention visualization of different heads.} To visualize a specific location's attention score, we sum the attention scores obtained for that location in all windows. Attention maps are up-sampled and overlaid on the image for better visualization.}
    \label{fig:attention_visualization}
\end{figure}

\begin{figure}
    \centering
    \includegraphics[width=0.95\linewidth]{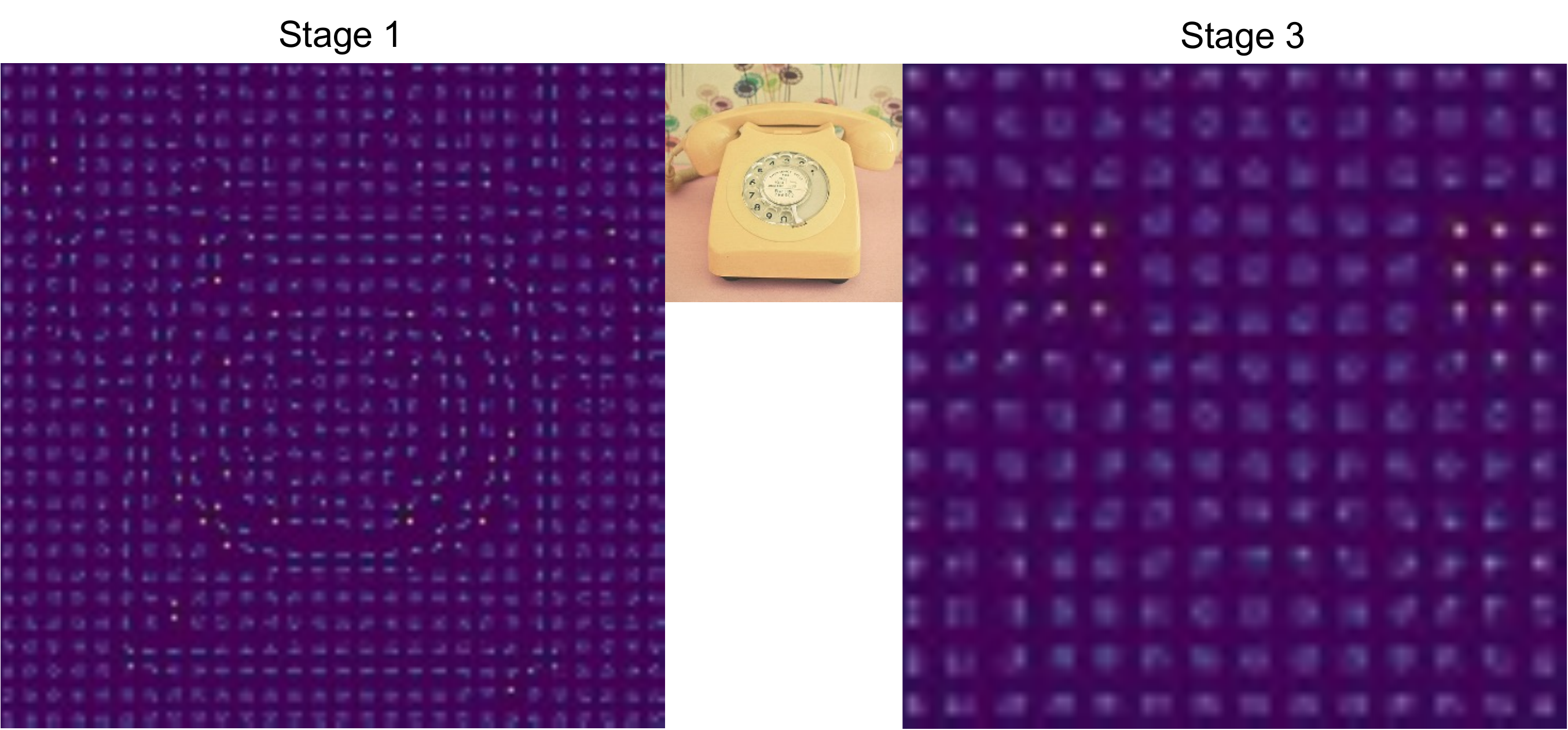}
    \caption{\textbf{QnA aggregation kernels visualization.} The attention kernels are tiled in the visualization instead of overlapped, causing the uniform grid effect. Brighter areas indicate higher attention scores. (best viewed when zoomed-in).}
    \label{fig:kernel_visualization}
\end{figure}

In QnA, the aggregation kernel of each window is derived from the attention scores between the learned queries and the window keys.  To visualize the attention of the whole image, we choose to sum the scores of each spatial-location as obtained in all relevant windows. You can find the visualization in~\Cref{fig:attention_visualization}. As shown in~\Cref{fig:attention_visualization}, the attentions are content-aware, suggesting the window aggregation kernels are spatially adaptive. For the learned kernels in each window, please refer to~\Cref{fig:kernel_visualization}.

\section{Full training details}
\label{sup:training_details}
\subsection{Image Classification}
We evaluate our method using the ImageNet-1K~\cite{ImageNet} benchmark, which contains 1.28M training images and 50,000 validation images from 1,000 classes. We follow the training recipe of DEiT~\cite{DEiT}, except we omit EMA~\cite{EMA} and repeated augmentations~\cite{RepeatedAug}. Particularly, we train all models for 300-epochs, using the AdmaW~\cite{Adam, AdamW} optimizer. We employ a linearly scaled learning rate $lr =\texttt{5e-4} \cdot \frac{\text{Batch size}}{256}$~\cite{LRScale}, with warmup epochs~\cite{WarmUp} varying according to model size and weight decay $wd = \texttt{5e-2}$. For augmentations, we apply RandAugment~\cite{RandAugment}, mixup~\cite{mixup} and CutMix~\cite{CutMix} with label-smoothing~\cite{LabelSmoothing}, and color-jitter~\cite{Jitter}. Finally, an increasing stochastic depth is applied~\cite{StochasticDepth}. Note this training recipe (with minor discrepancy between previous papers), is becoming the standard when training a Vision Transformer on the ImageNet benchmark. Finally, we normalize the queries in all QnA layers to be unit-vectors for better training stability.

\subsection{Object Detection}
We train the DETR model on COCO 2017 detection dataset~\cite{COCO}, containing $118K$ training images and a $5k$ validation set size. We utilize the training setting of DETR, in which the input is resized such that the shorter side is between 480 and 800 while the longer side is at most 1333~\cite{Detectron2}. An initial learning rate of $\mathtt{1e-4}$ is set for the detection transformer, and $\mathtt{1e-5}$ learning rate for the backbone network. Due to computational limitations, we use a short training scheduler of 75 epochs with a batch size of 32. The learning rates are scaled by 0.1 after 50 epochs. We trained DETR using the implementation in~\cite{Scenic}.
\section{QnA-ViT architecture}
\label{sup:arch}
The QnA-ViT architecture is composed of transformer blocks~\cite{ViT} and QnA blocks. First we split the image into $4 \times 4$ patches, and project them to form the input tokens. The vision transformer block is composed of a multihead attention layer (MSA) and an inverted-bottleneck feed forward network (FFN), with expansion 4. The output of block-$l$ is:
\begin{equation*}
\begin{split}
 &z'_l =  \mathbf{MSA}\left(\mathbf{LayerNorm}(z_{l-1})\right) + z_{l-1} \\
 &z_l = \mathbf{FFN}\left(\mathbf{LayerNorm}(z'_l)\right) + z'_l.
\end{split}
\end{equation*}
The QnA block shares a similar structure, except we replace the MSA layer with QnA layer. Downampling is performed using QnA blocks with stride set to 2 (to enable skip-connections we use $1\times 1$-convolution with similar stride). We employ pre-normalization~\cite{PreNorm} to stabilize training. Finally, we use global average pooling~\cite{GAP} right before the classification head, with LayerNorm~\cite{LayerNorm} employed prior to the pooling operation. Full architecture details can be found in~\Cref{tab:qna_vit_arch}.
\begin{table*}
\resizebox{\textwidth}{!}{%
\begin{tabular}{cc|cc|cc|cc}
\hline
Stage              & \multicolumn{1}{l|}{Output} & \multicolumn{2}{c|}{QnA-T}                                                                                   & \multicolumn{2}{c|}{QnA-S}                                                                                    & \multicolumn{2}{c}{QnA-B}                                                                                      \\ \cline{3-8} 
                   & \multicolumn{1}{l|}{}            & \multicolumn{1}{c|}{QnA Blocks}                               & SA Blocks                                    & \multicolumn{1}{c|}{QnA Blocks}                               & SA Blocks                                     & \multicolumn{1}{c|}{QnA Blocks}                               & SA Blocks                                      \\ \hline
\multirow{2}{*}{1} & \multirow{2}{*}{56x56}           & \multicolumn{2}{c|}{4x4 Conv, stride 4, dim 64}                                                              & \multicolumn{2}{c|}{4x4 Conv, stride 4, dim 64}                                                               & \multicolumn{2}{c}{4x4 Conv, stride 4, dim 96}                                                                 \\ \cline{3-8} 
                   &                                  & \multicolumn{1}{c|}{$\begin{bmatrix}\text{3x3 QnA-Block,}\\ \text{stride 1,}\\ \text{head 8}\end{bmatrix} \times 2$}  & None                                         &  \multicolumn{1}{c|}{$\begin{bmatrix}\text{3x3 QnA-Block,}\\ \text{stride 1,}\\ \text{head 8}\end{bmatrix} \times 2$}  & None                                          & \multicolumn{1}{c|}{$\begin{bmatrix}\text{3x3 QnA-Block,}\\ \text{stride 1,}\\ \text{head 6}\end{bmatrix} \times 2$}  & None                                           \\ \hline
\multirow{2}{*}{2} & \multirow{2}{*}{28x28}           & \multicolumn{2}{c|}{3x3 QnA, stride 2, head 16, dim 128}                                                      & \multicolumn{2}{c|}{3x3 QnA, stride 2, head 16, dim 128}                                                       & \multicolumn{2}{c}{3x3 QnA, stride 2, head 16, dim 192}                                                         \\ \cline{3-8} 
                   &                                  & \multicolumn{1}{c|}{$\begin{bmatrix}\text{3x3 QnA-Block,}\\ \text{stride 1,}\\ \text{head 16}\end{bmatrix} \times 3$} & None                                         & \multicolumn{1}{c|}{$\begin{bmatrix}\text{3x3 QnA-Block,}\\ \text{stride 1,}\\ \text{head 16}\end{bmatrix} \times 3$} & None                                          & \multicolumn{1}{c|}{$\begin{bmatrix}\text{3x3 QnA-Block,}\\ \text{stride 1,}\\ \text{head 12}\end{bmatrix} \times 3$} & None                                           \\ \hline
\multirow{2}{*}{3} & \multirow{2}{*}{14x14}           & \multicolumn{2}{c|}{3x3 QnA, stride 2, head 32, dim 256}                                                      & \multicolumn{2}{c|}{3x3 QnA, stride 2, head 32, dim 256}                                                       & \multicolumn{2}{c}{3x3 QnA, stride 2, head 32, dim 384}                                                         \\ \cline{3-8} 
                   &                                  & \multicolumn{1}{c|}{$\begin{bmatrix} \text{3x3 QnA-Block,}\\ \text{stride 1,}\\ \text{head 32} \end{bmatrix}\times 2$} & $\begin{bmatrix} \text{SA-Block,} \\ \text{win sz. 14 x 14,} \\ \text{head 8} \end{bmatrix} \times4$ & \multicolumn{1}{c|}{$\begin{bmatrix} \text{3x3 QnA-Block,}\\ \text{stride 1,}\\ \text{head 32} \end{bmatrix}\times 6$} & $\begin{bmatrix} \text{SA-Block,} \\ \text{win sz. 14 x 14,} \\ \text{head 8} \end{bmatrix} \times12$ & \multicolumn{1}{c|}{$\begin{bmatrix} \text{3x3 QnA-Block,}\\ \text{stride 1,}\\ \text{head 24} \end{bmatrix}\times 6$} &  $\begin{bmatrix} \text{SA-Block,} \\ \text{win sz. 14 x 14,} \\ \text{head 12} \end{bmatrix} \times12$ \\ \hline
\multirow{2}{*}{4} & \multirow{2}{*}{7x7}             & \multicolumn{2}{c|}{3x3 QnA, stride 2, head 64, dim 512}                                                      & \multicolumn{2}{c|}{3x3 QnA, stride 2, head 64, dim 512}                                                       & \multicolumn{2}{c}{3x3 QnA, stride 2, head 48, dim 768}                                                         \\ \cline{3-8} 
                   &                                  & \multicolumn{1}{c|}{None}                                     & $\begin{bmatrix} \text{SA-Block,}\\ \text{win sz. 7 x 7,}\\ \text{head 16} \end{bmatrix} \times2$  & \multicolumn{1}{c|}{None}                                     & $\begin{bmatrix} \text{SA-Block,}\\ \text{win sz. 7 x 7,}\\ \text{head 16} \end{bmatrix} \times2$    & \multicolumn{1}{c|}{None}                                     & $\begin{bmatrix} \text{SA-Block,}\\ \text{win sz. 7 x 7,}\\ \text{head 24} \end{bmatrix} \times2$     \\ \hline
\end{tabular}%
}
\caption{QnA-ViT architecture details. QnA is used to down-sample the feature maps between two consecutive stages. In stage 3 we first employ global self-attention blocks.}
\label{tab:qna_vit_arch}
\end{table*}
\section{Implementation \& complexity - extended}
\label{sup:implementation}
In this section we provide full details on the efficient implementation of QnA. To simplify the discussion, we only consider a single-query without positional embedding.

Let us first examine the output of a QnA layer, by expanding the softmax operation inside the attention layer:
\begin{equation*}
 \begin{split}
     z_{i,j}  &= \textbf{Attention}\left(\tilde{q},K_{\mathcal{N}_{i,j}}\right) \cdot V_{\mathcal{N}_{i,j}} \\ 
     & = \frac{\sum_{\mathcal{N}_{i,j}} e^{\tilde{q}K_{n,m}}v_{n,m} }{  \sum_{\mathcal{N}_{i,j}} e^{\tilde{q}K_{n,m}}}.
 \end{split}
\end{equation*}

Recall, $\mathcal{N}_{i,j}$ is the $k\times k$-window at location $(i,j)$. While it may seem that we need to calculate the query-key dot product for each window, notice that since we use the same query over each window, then we can calculate $\tilde{q}K^T$ once for the entire input. Also, we can leverage the matrix multiplication associativity and improve the computation complexity by calculating $\tilde{q} W_k^T$ first (this fused implementation reduces the memory by avoiding the allocation of the key entities). Once we calculate the query-key dot product, we can efficiently aggregate the dot products using the sum-reduce operation supported in many deep learning frameworks (e.g., Jax~\cite{jax2018github}). More specifically, let $\mathbf{Sum}_k(\dots)$ be a function that sums-up values in each $k \times k$-window,  then:
$$z_{i,j} = \frac{\sum_{\mathcal{N}_{i,j}} e^{\tilde{q}K_{n,m}}v_{n,m} }{  \sum_{\mathcal{N}_{i,j}} e^{\tilde{q}K_{n,m}}} = \frac{\mathbf{Sum}_k \left(e^{\tilde{q}K^T}*V \right)} {\mathbf{Sum}_k \left(e^{\tilde{q}K^T} \right)},$$
where * is the element wise multiplication. A pseudo-code of our method can be found in~\Cref{alg:qna}. We further provide a code-snippet of the QnA module, implemented in Jax/Flax~\cite{jax, flax} (see~\Cref{fig:qna_jax}).

\paragraph{Complexity Analysis: } in the single query variant, extracting the values and computing the key-query dot product require $2HWD^2$ computation and $HW + HWD$ extra space. Additionally, computing the softmax using the above method requires additional $\mathcal{O}(k^2HWD)$ computation (for summation and division), and $\mathcal{O}(HWD)$ space (which is independent of $k$, i.e., the window size). For multiple-queries variant ($L=2$), see empirical comparison in~\cref{fig:complexity_comparison}.

\definecolor{codegreen}{rgb}{0,0.6,0}
\definecolor{codegray}{rgb}{0.5,0.5,0.5}
\definecolor{codepurple}{rgb}{0.58,0,0.82}
\definecolor{backcolour}{rgb}{0.95,0.95,0.92}

\lstdefinestyle{mystyle}{
    backgroundcolor=\color{backcolour},   
    commentstyle=\color{codegreen},
    keywordstyle=\color{magenta},
    numberstyle=\tiny\color{codegray},
    stringstyle=\color{codepurple},
    basicstyle=\ttfamily\footnotesize,
    breakatwhitespace=false,         
    breaklines=true,                 
    captionpos=b,                    
    keepspaces=true,                 
    numbers=left,                    
    numbersep=5pt,                  
    showspaces=false,                
    showstringspaces=false,
    showtabs=false,                  
    tabsize=2
}

\lstset{style=mystyle}

\begin{figure*}
    \centering
\begin{lstlisting}[language=Python]
class QnA(nn.Module):
  @nn.compact
  def __call__(self, X):
    # Initialize Parameters:
    Q  = # Query vectors  [L, h, D//h]
    Wk = # Linear Projection for keys [D, D]
    Ws = # Attention weight scale [k, k, L * h]
    B_rpe = #Relative PE [k, k, L * h]
    # Fused implementation of Q*(X*W_K).Transpose 
    Wk = Wk.reshape([-1, heads, D // heads])
    QWk = jnp.einsum('lhd,Dhd->Dlh', Q, Wk)  
    QK_similariy = jnp.einsum('BHWD,BDqh->BHWqh', X, QWk)
    # Compute V
    V = nn.Dense(D)(X).reshape([B, H, W, heads, D // heads])
    # Compute Attention:
    exp_similarity = jnp.exp(QK_similariy)  # [B, H, W, L, h]
    exp_similarity_v = exp_similarity[..., jnp.newaxis] * V[:,:,: jnp.newaxis,:]  # [B, H, W, L, D]
    aux_kernel = (jnp.exp(B_rpe) * Ws).repeat(repeats=D // heads, axis=-1)  # [k, k, d, L*h]
    A = jax.lax.conv_general_dilated(exp_similarity_v.reshape([B, H, W, LD]),
                            aux_kernel, window_strides=[s, s], padding='SAME',
                            feature_group_count=L * D,).reshape([B, H_out, W_out, L, heads, D // heads])
    A = jnp.reshape(A, [B, H_out, W_out, L, heads, D // heads])
    aux_kernel = jnp.exp(B_rpe)  # [k, k, 1, L*h]
    B = jax.lax.conv_general_dilated(exp_similarity.reshape([B, H, W, -1]),
                                     aux_kernel,
                                     window_strides=[s, s],
                                     padding='SAME',
                                     dimension_numbers=_conv_dimension_numbers(I.shape)
                                     ).reshape([B, H_out, W_out, L, heads, 1])
    out_heads = jnp.sum(A / B, axis=-2).reshape([B, H_out, W_out, D])
    final_out = nn.Dense(self.D)(out_heads)
    return final_out
\end{lstlisting}
    \caption{Code snippet of the QnA module implemented in Jax~\cite{jax} and Flax~\cite{flax}. Full implementation, with pre-trained networks weights, will be made publicly available.}
    \label{fig:qna_jax}
\end{figure*}

\renewcommand{\algorithmicrequire}{\textbf{Input:}}
\renewcommand{\algorithmicensure}{\textbf{Parameters:}}
\algnewcommand{\LineComment}[1]{\Statex \(\textbf{//}\) \textit{#1}}

\algdef{SE}[SUBALG]{Indent}{EndIndent}{}{\algorithmicend\ }%
\algtext*{Indent}
\algtext*{EndIndent}

\begin{algorithm}
\caption{Efficient implementation of QnA. All operations can be implemented efficiently with little memory-overhead. Further, $\mathbf{Sum}_k$ applies sum-reduction to all elements in each $k\times k$-window.}\label{alg:qna}
\begin{algorithmic}[1]
\Require $X \in \mathbb{R}^{H\times W \times D}$
\Ensure  $W_K, W_V \in \mathbb{R}^{D \times D}$, $\tilde{q} \in \mathbb{R}^{D}$
\LineComment{Compute values:}
\Indent
    \State $V \gets XW_V$
\EndIndent
\LineComment{Compute the query-key dot product ( $\mathbf{S \gets \tilde{q}K^T}$):}
\Indent
    \State Let $A \gets \tilde{q}W_K^T$
    \For{$l \in \left[ L\right], i,j \in \left[ H \right] \times \left[W \right]$}
    \State $S_{l,i,j} \gets A_{l,i,j}\cdot X_{i,j}^T$
    \EndFor
\EndIndent
\LineComment{Compute the final output:}
\Indent
    \State Let $B \gets e^{S}$ be the element-wise exponent of $S$
    \State Let $C \gets B*V$ \Comment{$*$ is the element-wise product}
    \State \Return $\mathbf{Sum}_k(C) / \mathbf{Sum}_k(B)$
\EndIndent
\end{algorithmic}
\end{algorithm}
\section{Design choices - full report}
\label{sup:design_choices}
\subsection{Number of queries}
\label{sup:num_of_queries}
To verify the effectiveness of using multiple queries, we trained a lightweight QnA-ViT network composed of \textit{local self-attention blocks} and \textit{QnA blocks}. We set the window size of all local self-attention layers to be 7x7, and we use a 3x3 receptive field for QnA layers. All the downsampling performed are QnA-Based. A similar architecture was used for the SASA~\cite{SASA} baseline, where we replaced the QnA layers with SASA layers. The number of QnA/SASA blocks used for each stage are ${[}2,2,2,0 {]}$ and the number of local self-attention blocks are ${[}1,1,5,2{]}$.

\subsection{Number of heads}
\label{sup:num_of_heads}
Using more attention heads is beneficial for QnA. More specifically, we conducted two experiments, one where we use only QnA blocks and the second where we use both QnA blocks and (global) self-attention blocks. In the first experiment, we use the standard ImageNet training preprocessing~\cite{Inception}, meaning we employ random crop with resize and random horizontal flip. In the second experiment, we used DeiT training preprocessing. We show the full report in~\Cref{tab:heads_qna_only} and~\Cref{tab:heads_qna_and_sa}. 

First, from~\Cref{tab:heads_qna_only}, we notice that training shallow QnA-networks for fewer epochs requires many attention heads. Furthermore, it is better to maintain a fixed dimension head across stages - this is done by doubling the number of heads between two consecutive stages. For deeper networks, the advantage of using more heads becomes less significant. This is because the network can capture more feature subspaces by leveraging its additional layers. Finally, when using both QnA and ViT blocks, it is still best to use more heads for QnA layers, while for ViT blocks, it is best to use a high dimension representation by having fewer heads (see~\Cref{tab:heads_qna_and_sa}).

\begin{table}
\resizebox{\linewidth}{!}{%
\begin{tabular}{@{}clccc@{}}
\toprule[1.5pt]
QnA Blocks                     & Heads            & AugReg                & Epochs                                   & Top-1 Acc. \\ \midrule[1.25pt]
\multirow{6}{*}{{[}2,2,2,2{]}} & {[}2,2,2,2{]}    & \multirow{6}{*}{None} & \multicolumn{1}{c|}{\multirow{6}{*}{90}} & 68.33      \\
                               & {[}4,4,4,4{]}    &                       & \multicolumn{1}{c|}{}                    & 69.05      \\
                               & {[}8,8,8,8{]}    &                       & \multicolumn{1}{c|}{}                    & 69.98      \\
                               & {[}2,4,8,16{]}   &                       & \multicolumn{1}{c|}{}                    & 70.08      \\
                               & {[}4,8,16,32{]}  &                       & \multicolumn{1}{c|}{}                    & 70.54      \\
                               & {[}8,16,32,64{]} &                       & \multicolumn{1}{c|}{}                    & \textbf{71.12}      \\ \midrule
\multirow{3}{*}{{[}3,4,6,3{]}} & {[}2,4,8,16{]}   & \multirow{3}{*}{None} & \multicolumn{1}{c|}{\multirow{3}{*}{90}} & 72.66      \\
                               & {[}4,8,16,32{]}  &                       & \multicolumn{1}{c|}{}                    & 72.82      \\
                               & {[}8,16,32,64{]} &                       & \multicolumn{1}{c|}{}                    & \textbf{73.02}      \\ \bottomrule[1.5pt]
\end{tabular}%
}
\caption{\textbf{The affect of number of heads on QnA.} We train two networks using the Inception preprocessing~\cite{Inception}, i.e., random crop and horizontal flip. We set the number of QnA blocks according to the ResNet-18 and ResNet-50 networks (the number of blocks for each stage is stated in the first column). As can be seen, using fixed head-dimension is better than increasing the head-dimension as we propagate through the network. Shallow networks benefit from having many heads, while deeper networks gain less from more heads. Therefore we suggest increasing the head-dimension for deeper networks for better memory utilization.}
\label{tab:heads_qna_only}
\end{table}
\begin{table}
\resizebox{\linewidth}{!}{%
\begin{tabular}{@{}llllcc|c@{}}
\toprule[1.5pt]
\multicolumn{2}{c}{Blocks}                                      & \multicolumn{2}{c}{Heads}           & \multicolumn{1}{l}{\multirow{2}{*}{AugReg}} & \multicolumn{1}{l|}{\multirow{2}{*}{Epochs}} & \multirow{2}{*}{Top-1 Acc.} \\
QnA                            & SA                             & QnA              & SA               & \multicolumn{1}{l}{}                        & \multicolumn{1}{l|}{}                        &                             \\ \midrule[1.25pt]
\multirow{5}{*}{{[}1,2,3,0{]}} & \multirow{5}{*}{{[}1,1,3,2{]}} & {[}2,4,8,16{]}   & {[}2,4,8,16{]}   & \multirow{5}{*}{DeiT~\cite{DEiT}}                       & \multirow{5}{*}{90}                          & 78.17                       \\
                               &                                & {[}4,8,16,32{]}  & {[}4,8,16,32{]}  &                                             &                                              & 79.24                       \\
                               &                                & {[}8,16,32,64{]} & {[}8,16,32,64{]} &                                             &                                              & 79.34                       \\
                               &                                & {[}8,16,32,64{]} & {[}2,4,8,16{]}   &                                             &                                              & 79.31                       \\
                               &                                & {[}8,16,32,64{]} & {[}4,8,16,32{]}  &                                             &                                              & \textbf{79.53}                       \\ \midrule[1.25pt]
\multirow{2}{*}{{[}1,2,3,0{]}} & \multirow{2}{*}{{[}1,1,3,2{]}} & {[}8,16,32,64{]} & {[}2,4,8,16{]}   & \multirow{2}{*}{DeiT~\cite{DEiT}}                       & \multirow{2}{*}{300}                         & \textbf{81.5}                        \\
                               &                                & {[}8,16,32,64{]} & {[}4,8,16,32{]}  &                                             &                                              & 81.49                       \\ \bottomrule[1.5pt]
\end{tabular}%
}
\caption{\textbf{The number of heads affect on QnA and ViT Blocks.} QnA still benefits from more heads, while ViT blocks need higher-dimension representation, specially for longer training.}
\label{tab:heads_qna_and_sa}
\end{table}

\subsection{How many QnA layers do you need?} 
\label{sup:how_much_qna}
To understand the benefit of using QnA layers, we consider a dozen network architectures that combine vanilla ViT and QnA blocks. For ViT blocks, we tried to use global attention in the early stages but found it better to use local self-attention and restrict the window size to be at most 14x14. We group the architecture choices into three groups:

\begin{enumerate}
    \item We consider varying the number of QnA blocks in the early stages.
    \item We change the number of QnA blocks in the third stage.
    \item We use lower window size for the local-self attention blocks. Namely, 7x7 window in all ViT blocks at all stages.
\end{enumerate}
Finally, all networks were trained for 300 epochs following DeiT preprocessing. The full report can be found in~\Cref{tab:how_much_qna}. 

From~\Cref{tab:how_much_qna}, local-self attention is not very beneficial in the early stages and can be omitted by using QnA blocks only. Furthermore, using more global-attention blocks in deeper stages is better, but the network’s latency can be reduced by having a considerable amount of QnA blocks. Finally, local self-attention becomes less effective when using a lower window size. In particular, since QnA is shift-invariant, it can mitigate the lack of cross-window interactions, reflected in the improvement gain we achieve when using more QnA blocks.

\begin{table}
\centering
\resizebox{.9\linewidth}{!}{%
\begin{tabular}{llccc}
\toprule[1.5pt]
\multicolumn{2}{c}{Blocks}                       & \multirow{2}{*}{Params} & \multirow{2}{*}{GFLOPS} & \multirow{2}{*}{Top- Acc} \\
\multicolumn{1}{c}{QnA} & \multicolumn{1}{c}{SA} &                         &                         &                           \\ \midrule[1.25pt]
\multicolumn{5}{c}{Changes in stages 1, 2}                                                                           \\ \midrule[1.25pt]
{[}1,1,4,0{]}           & {[}3,3,3,2{]}          & 16.62M                  & 3.200                   & 81.70                     \\ \hline
{[}2,2,4,0{]}           & {[}2,2,3,2{]}          & 16.51M                  & 2.909                   & 81.74                     \\ \hline
{[}3,3,4,0{]}           & {[}1,1,3,2{]}          & 16.40M                  & 2.875                   & \textbf{81.86}            \\ \hline
{[}4,4,4,0{]}           & {[}0,0,3,2{]}          & \textbf{16.30M}         & \textbf{2.584}          & 81.83                     \\ \midrule[1.25pt]
\multicolumn{5}{c}{Changes in stage 3}                                                                               \\ \midrule[1.25pt]
{[}4,4,7,0{]}           & {[}0,0,0,2{]}          & 16.00M                  & 2.631                   & 80.8                      \\ \hline
{[}4,4,5,0{]}           & {[}0,0,2,2{]}          & 16.25M                  & 2.698                   & 81.30                     \\ \hline
{[}4,4,3,0{]}           & {[}0,0,4,2{]}          & \textbf{16.40M}         & \textbf{2.628}          & \textbf{81.9}             \\ \hline
{[}4,4,1,0{]}           & {[}0,0,6,2{]}          & 16.55M                  & 2.714                   & \textbf{81.9}             \\ \midrule[1.25pt]
\multicolumn{5}{c}{Local SA with window size $7\times7$}                                                                    \\ \midrule[1.25pt]
{[}1,1,1,0{]}           & {[}2,2,6,2{]}          & 16.38M                  & 2.568                   & 80.0                      \\ \hline
{[}2,2,2,0{]}           & {[}1,1,5,2{]}          & 16.3M                   & 2.491                   & 80.6                      \\ \hline
{[}2,2,3,0{]}           & {[}1,1,4 ,2{]}         & 16.23M                  & 2.471                   & 80.7                      \\ \hline
{[}2,2,4,0{]}           & {[}1,1,3, 2{]}         & \textbf{16.16M}         & \textbf{2.450}          & \textbf{80.8}             \\ \bottomrule[1.5pt]
\end{tabular}%
}
\caption{\textbf{How much QnA do you really need? - full report.} The number of QnA and local self-attention (SA) blocks in each stage are indicated in the first row. In the first two sub-tables, a window size of $14\times 14$ except in the last stage, where a $7\times 7$ window size was set. In the last sub-table, we reduce the window size to become $7\times 7$ for all stages.}
\label{tab:how_much_qna}
\end{table}
\section{QnA Variants - extended version}
\label{sup:qna_variants}
As discussed in the paper, we incorporate multi-head attention and positional embedding in our layer. 

\paragraph{Positional Embedding} Self-attention is a permutation invariant operation, meaning it does not assume any spatial relations between the input tokens. This property is not desirable in image processing, where relative context is essential. Position encoding can be injected into the self-attention mechanism to solve this. Following recent literature, we use relative-positional embedding~\cite{RPE1, RPE2, RPE3, RPE4, RethinkingPE}. This introduces a spatial bias into the attention scheme, rendering~\cref{eq:self-attention} (from the main text) now to be:

\begin{equation}
\label{eq:attention_with_relative_pe}
    \textbf{Attention}\left(Q, K\right) = \textbf{Softmax}\left(QK^{T} / {\sqrt{D}} + B \right), 
\end{equation} where $B \in \mathbb{R}^{k \times k}$ is a learned relative positional encoding. Note, different biases are learned for each query in the QnA layer, which adds $\mathcal{O}(L\times k^2)$ additional space.

\paragraph{Multi-head attention:} As in the original self-attention layer~\cite{AttentionIsAlYouNeed}, we use multiple heads in order to allow the QnA layer to capture different features simultaneously. In fact, as we will show in~\Cref{sup:num_of_heads}, using more attention heads is beneficial to QnA. 

In mutli-head attention, all queries $Q$, keys $K$, and values $V$ entities are split into $h$ sub-tensors, which will correspond to vectors in the lower dimensional space $\mathbb{R}^{D/h}$. More specifically, let $Q^{(i)},K^{(i)},V^{(i)}$ be the $i$-th sub vector of each query, key and value, respectively. The self-attention of head-$i$ becomes: 

\begin{equation}
\label{eq:multi_head_attention_with_relative_pe}
\begin{split}
    \text{head}_i &= \textbf{Attention}\left(Q^{(i)}, K^{(i)}\right)V^{(i)} \\
    & = \textbf{Softmax}\left(Q^{(i)}K^{(i)T} / {\sqrt{d_h}} + B \right) V^{(i)},
\end{split}
\end{equation}, where $d_{h} = D/h $, and the output of the Multi-Head Attention (MHA) is:

\begin{equation}
\label{eq:final_multihead_attention_with_relative_pe}
\begin{split}
    \textbf{MHA} & \left( Q,K,V \right) = \\
    & \texttt{Concat}\left( \text{head}_1,  \dots, \text{head}_h \right)W_O
\end{split}
\end{equation}, where $W_O \in \mathbb{R}^{D\times D}$ is the final projection matrix.

\paragraph{Up-sampling using QnA} Up-sampling by factor $s$ can be defined using $s^2$-learned queries, i.e., $\tilde{Q} \in \mathbb{R}^{s^2\times D}$. To be precise, the output of each window $\mathcal{N}_{i,j}$ is expressed via:
\begin{equation}
\label{eq:s_qna}
    z_{i,j}  = \textbf{Attention}\left(\tilde{Q},K_{\mathcal{N}_{i,j}}\right) \cdot V_{\mathcal{N}_{i,j}}.
\end{equation}
Note, the window output given by \cref{eq:s_qna} is now a matrix of size $s^2 \times D$, and the total output $Z$ is a tensor of size $H \times W \times s^2 \times D$. To form the up-sampled output $Z^{s}$, we need to reshape the tensor and permute its axes:

\begin{equation}
\label{eq:upsample_qna}
\begin{split}
    Z^{s}_{(1)} &  = \text{Reshape}(Z,  [ H, W, s, s, D ] ) \\
    Z^{s}_{(2)} & = \text{Permute}(Z^{s}_{(1)},  [  0, 2, 1, 3, 4]) \\
    Z^{s} & = \text{Reshape}(Z^{s}_{(2)},  [H\times s, W\times s, D]) .    
\end{split}
\end{equation}
\section{QnA as an upsampling layer}
\label{sup:auto-encoders}
In the paper, we showed how QnA could be used as an upsampling layer. In particular, we trained a simple auto-encoder network composed of five downsampling layers and five upsampling layers. We use the $L_1$ reconstruction loss as an objective function to train the auto-encoder. We considered three different encoder layers:
\begin{itemize}
    \item \textbf{ConvS2-IN}: 3x3 convolution with stride 2 followed by an Instance Normalization layer~\cite{InstanceNorm}.
    \item \textbf{Conv-IN-Max}: 3x3 convolution with stride 1 followed by an Instance Normalization layer and max-pooling with stride 2.
    \item \textbf{LN-QnA}: Layer Normalization~\cite{LayerNorm} followed by a 3x3 single-query QnA layer (without skip-connections).
\end{itemize}

and three different decoder layers:

\begin{itemize}
    \item \textbf{Bilinear-Conv-IN}: x2 bilinear up-sampling followed by 3x3 convolution and Instance Normalization.
    \item \textbf{ConvTransposed-IN}: A 2d transposed convolution followed by Instance Normalization.
    \item \textbf{LN-UQnA}: Layer Normalization followed by a 3x3 up-sampling QnA layer (without skip-connections)
\end{itemize}

For our baseline networks, we found it best to use \textbf{Conv-In-Max} in the encoder path, and chose either \textbf{Bilinear-Conv-In} or \textbf{ConvTranspose-IN} for the decoder path. For QnA-based auto-encoders, we use QnA layers for both down-sampling and up-sampling. All networks are trained on the CelebA dataset~\cite{CelebA}, where all images are center-aligned and resized to resolution of size $256^2$. All networks were trained for 10-epochs, using the Adam~\cite{Adam} optimizer (learning rates were chosen according to the best test-loss).

\end{document}